%% file: bare_jrnl.tex
\documentclass[journal]{IEEEtran}
\usepackage{cite}
\usepackage{amsmath,amssymb,amsfonts}
\usepackage{algorithmicx}
\usepackage{algorithm}
\usepackage{algpseudocode}
\usepackage{graphicx}
\usepackage{textcomp}
\usepackage{amsmath,amsfonts}
\usepackage{array}
\usepackage[caption=false,font=normalsize,labelfont=sf,textfont=sf]{subfig}
\usepackage{textcomp}
\usepackage{stfloats}
\usepackage{url}
\usepackage{verbatim}
\usepackage{graphicx}
\usepackage{cite}
\usepackage{graphicx}
\usepackage{amsmath}
\usepackage{amssymb}
\usepackage{booktabs}
\usepackage{array,multirow,graphicx}
\usepackage{pifont}
\usepackage{setspace}
\usepackage{bbm}
\usepackage{caption}
\usepackage{comment}
\usepackage{tabularx}
\usepackage{booktabs}

\usepackage{hyperref}
\usepackage[dvipsnames]{xcolor}

\usepackage{xcolor}
\usepackage[normalem]{ulem}

\ifCLASSINFOpdf
\else
\fi

\begin{document}

\title{Global Collinearity-aware Polygonizer for Polygonal Building Mapping in Remote Sensing}

\author{Fahong Zhang, Yilei Shi,~\IEEEmembership{Member,~IEEE,}
and Xiao Xiang Zhu,~\IEEEmembership{Fellow,~IEEE,}
\thanks{The work is jointly supported by the German Research Foundation (DFG GZ: ZH 498/18-1; Project number: 519016653), by the German Federal Ministry of Education and Research (BMBF) in the framework of the international future AI lab "AI4EO -- Artificial Intelligence for Earth Observation: Reasoning, Uncertainties, Ethics and Beyond" (grant number: 01DD20001), by German Federal Ministry for Economic Affairs and Climate Action in the framework of the "national center of excellence ML4Earth" (grant number: 50EE2201C), by the German Federal Ministry for the Environment, Nature Conservation, Nuclear Safety and Consumer Protection (BMUV) based on a resolution of the German Bundestag (grant number: 67KI32002B; Acronym: \textit{EKAPEx}), and by the Munich Center for Machine Learning.}
\thanks{(Correspondence: Xiao Xiang Zhu)}
\thanks{
F. Zhang, X. Zhu are with the Chair of Data Science in Earth Observation, Technical University of Munich, 80333 Munich, Germany (e-mail: (fahong.zhang, xiaoxiang.zhu)@tum.de).
}
\thanks{
X. Zhu is also with the Munich Center for Machine Learning.
}
\thanks{Y. Shi is with School of Engineering and Design, Technical University of Munich (TUM), 80333 Munich, Germany
(e-mail: yilei.shi@tum.de)}}

\markboth{IEEE Transactions on Geoscience and Remote Sensing ,~Vol.~xx, No.~xx, March~2025}%
{Shell \MakeLowercase{\textit{et al.}}: Bare Demo of IEEEtran.cls for Journals}

\maketitle

\begin{abstract}
\input{content/abstract.tex}
\end{abstract}


\section{Introduction}
\label{sec:introduction}
\input{content/introduction}

\section{Related Works}
\input{content/related_works}

\section{Methods}
\input{content/methods}

\section{Experiments}
\input{content/experiments}

\section{Conclusion}
\input{content/conclusion}


\bibliographystyle{IEEEtran}
\bibliography{mybib}

\ifCLASSOPTIONcaptionsoff
  \newpage
\fi
%

\begin{IEEEbiography}[{\includegraphics[width=1in,height=1.25in,clip,keepaspectratio]{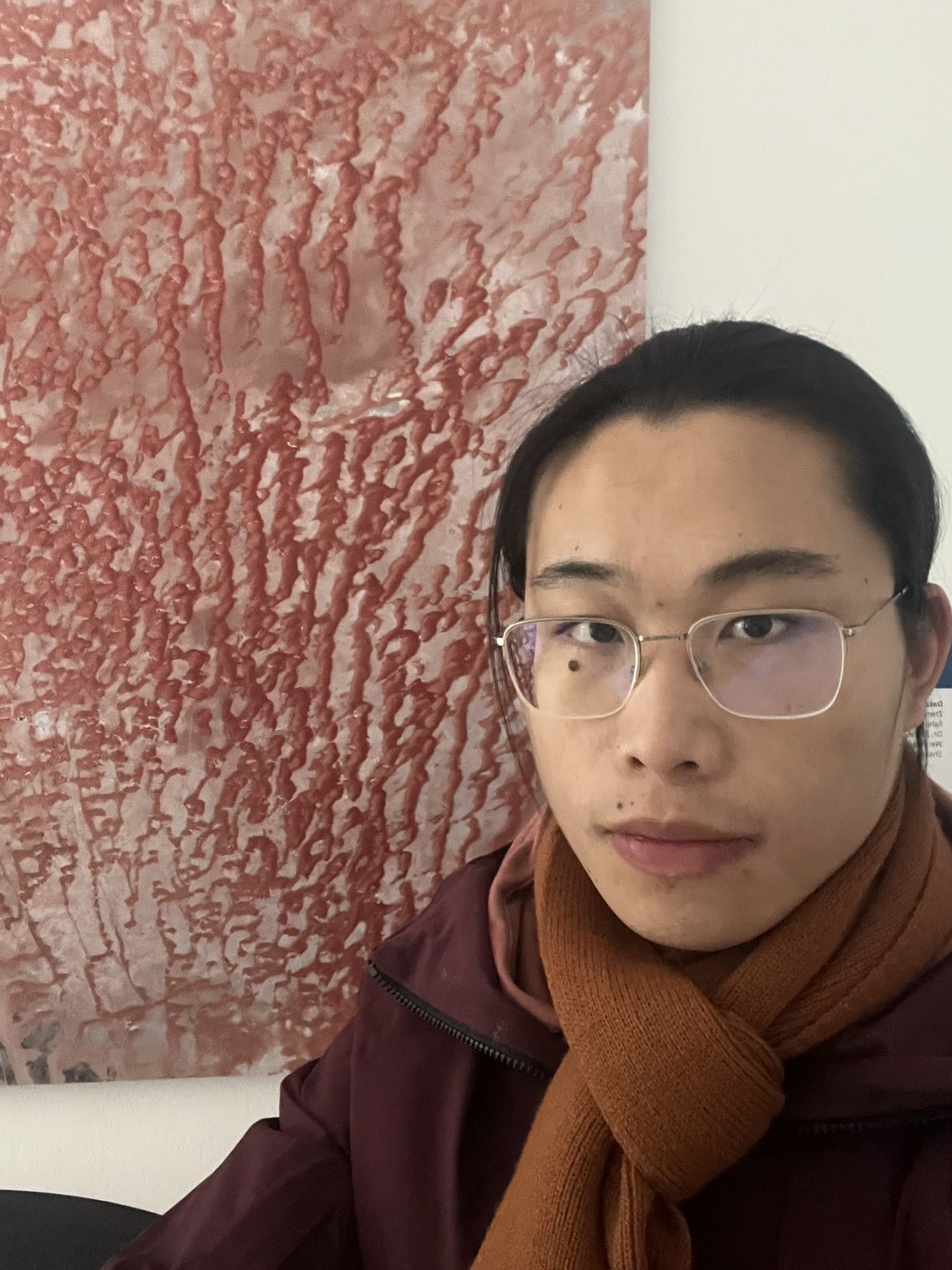}}]
{Fahong Zhang} received the B.E. degree in software engineering from Northwestern Polytechnical University, Xi'an, China, in 2017,
the M.S. degree in computer science with the Center for OPTical IMagery Analysis and Learning, Northwestern Polytechnical University, Xi'an, China.
He is now pursuing the Ph.D. degree with the department of Aerospace and Geodesy, Data Science in Earth Observation, Technical University of Munich.
His reserach interests include computer vision and satellite image processing.
\end{IEEEbiography}

\begin{IEEEbiography}[{\includegraphics[width=1in,height=1.25in,clip,keepaspectratio]{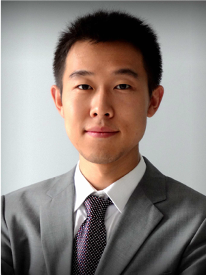}}]
{Yilei Shi} (Member, IEEE) received the Dipl.-Ing.
degree in mechanical engineering and the Dr.-Ing.
degree in signal processing from the Technical University of Munich (TUM), Munich, Germany, in
2010 and 2019, respectively.
He is a Senior Scientist with the Chair of Remote
Sensing Technology, TUM. His research interests
include fast solver and parallel computing for large-scale problems, high-performance computing and
computational intelligence, advanced methods on
synthetic aperture radar (SAR) and InSAR processing, machine learning, and deep learning for variety of data sources, such
as SAR, optical images, and medical images, and partial differential equation (PDE)-related numerical modeling and computing.
\end{IEEEbiography}

\begin{IEEEbiography}[{\includegraphics[width=1in,height=1.25in,clip,keepaspectratio]{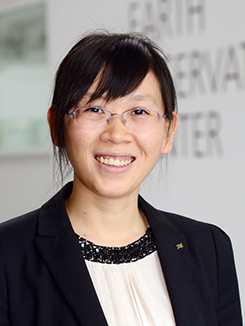}}]{Xiao Xiang Zhu}(S'10--M'12--SM'14--F'21) received the Master (M.Sc.) degree, her doctor of engineering (Dr.-Ing.) degree and her “Habilitation” in the field of signal processing from Technical University of Munich (TUM), Munich, Germany, in 2008, 2011 and 2013, respectively.
\par
She is the Chair Professor for Data Science in Earth Observation at Technical University of Munich (TUM) and was the Founding Head of the Department ``EO Data Science'' at the Remote Sensing Technology Institute, German Aerospace Center (DLR). Since 2019, Zhu is a co-coordinator of the Munich Data Science Research School (www.mu-ds.de). Since 2019 She also heads the Helmholtz Artificial Intelligence -- Research Field ``Aeronautics, Space and Transport". Since May 2020, she is the PI and director of the international future AI lab "AI4EO -- Artificial Intelligence for Earth Observation: Reasoning, Uncertainties, Ethics and Beyond", Munich, Germany. Since October 2020, she also serves as a Director of the Munich Data Science Institute (MDSI), TUM. Prof. Zhu was a guest scientist or visiting professor at the Italian National Research Council (CNR-IREA), Naples, Italy, Fudan University, Shanghai, China, the University  of Tokyo, Tokyo, Japan and University of California, Los Angeles, United States in 2009, 2014, 2015 and 2016, respectively. She is currently a visiting AI professor at ESA's Phi-lab. Her main research interests are remote sensing and Earth observation, signal processing, machine learning and data science, with their applications in tackling societal grand challenges, e.g. Global Urbanization, UN’s SDGs and Climate Change.

Dr. Zhu is a member of young academy (Junge Akademie/Junges Kolleg) at the Berlin-Brandenburg Academy of Sciences and Humanities and the German National  Academy of Sciences Leopoldina and the Bavarian Academy of Sciences and Humanities. She serves in the scientific advisory board in several research organizations, among others the German Research Center for Geosciences (GFZ) and Potsdam Institute for Climate Impact Research (PIK). She is an associate Editor of IEEE Transactions on Geoscience and Remote Sensing and serves as the area editor responsible for special issues of IEEE Signal Processing Magazine. She is a Fellow of IEEE.

\end{IEEEbiography}




\end{document}

%% file: content/abstract.tex
This paper addresses the challenge of mapping polygonal buildings from remote sensing images and introduces a novel algorithm, the Global Collinearity-aware Polygonizer (GCP).
GCP, built upon an instance segmentation framework, processes binary masks produced by any instance segmentation model. 
The algorithm begins by collecting polylines sampled along the contours of the binary masks. These polylines undergo a refinement process using a transformer-based regression module to ensure they accurately fit the contours of the targeted building instances.
Subsequently, a collinearity-aware polygon simplification module simplifies these refined polylines and generate the final polygon representation.
This module employs dynamic programming technique to optimize an objective function that balances the simplicity and fidelity of the polygons, achieving globally optimal solutions. Furthermore, the optimized collinearity-aware objective is seamlessly integrated into network training, enhancing the cohesiveness of the entire pipeline.
The effectiveness of GCP has been validated on two public benchmarks for polygonal building mapping. Further experiments reveal that applying the collinearity-aware polygon simplification module to arbitrary polylines, without prior knowledge, enhances accuracy over traditional methods such as the Douglas-Peucker algorithm. This finding underscores the broad applicability of GCP.
The code for the proposed method will be made available at \url{https://github.com/zhu-xlab}.

%% file: content/introduction.tex
Extracting building footprints from high-resolution remote sensing images is crucial for understanding the human urbanization process \cite{zhu2024global} and serves as a foundational task for applications such as population estimation \cite{robinson2017deep} and flood risk management \cite{rentschler2022flood}. While extensive research has focused on generating raster-format building footprints as binary masks \cite{guo2024high}, there remains a significant demand for vector formats. Vector formats offer more structured and compact representations, ideal for the construction of digital twins \cite{jiang2021industrial} and can provide a foundation for developing 3D building models  \cite{chen2021mask}.

Building on prior research, several effective paradigms for polygonal building mapping have been introduced.
One straightforward idea is the mask-based methods \cite{xu2023hisup,girard2021polygonal}, which generate initial building polygons by processing predicted binary masks from a semantic segmentation model using the Marching Squares \cite{lorensen1998marching} or contour tracing \cite{suzuki1985topological} algorithms. These initial polygons are then simplified through polyline simplification techniques, such as the Douglas-Peucker algorithm \cite{douglas1973algorithms}, to refine their structure and reduce complexity. 
However, this pipeline encounters specific challenges.
First, binary masks generated by semantic segmentation models based on Convolutional Neural Networks (CNNs) \cite{krizhevsky2012imagenet} often exhibit curved or rounded edges at the corners of building polygons. This characteristic can lead to inaccuracies in the final polygon shapes when converting from binary masks to polygonal representations.
Second, when simplifying an initial polygon that has excessively sampled vertices around the building's contour, there is a notable performance decay due to the ineffectiveness of the polygon simplification method.

To address the ambiguity in binary masks at edges and corners, contour-based methods \cite{hu2023polybuilding,zhang2024hit} have been developed. These methods typically adopt a workflow similar to object detection  \cite{carion2020end} pipelines but, instead of predicting a bounding box, they predict a fixed number of points around the building's contour in a specified order, either clockwise or anti-clockwise. To generate the final polygon, these methods often incorporate a classification layer for each predicted vertex to filter out redundant vertices with low classification scores, complemented by a non-maximum suppression (NMS) layer to further refine the vertex selection.
The primary limitation of this method is its inability to accurately represent complex building structures that contain internal holes, as it predicts only a single contour for each building. Additionally, the approach of filtering vertices based on classification scores followed by NMS may fail to generalize well when the contour predictions exhibit high uncertainty.

Another technique, termed the connection-based method, initiates by predicting potential primitives (e.g., vertices and edges) for buildings across the entire image. It then employs a solver for the linear sum assignment problem, such as the Sinkhorn algorithm \cite{cuturi2013sinkhorn}, to organize these primitives into polygons based on their pairwise connection tendencies. However, as noted in the literature \cite{zorzi2022polyworld}, this method encounters challenges when two adjacent buildings are very close together, often resulting in shared vertices or other primitives. This poses a problem as each primitive can only be assigned to one polygon, potentially leading to inaccuracies in polygon formation.

This paper centers on the straightforward idea of converting masks back into polygons. We introduce the Global Collinearity-aware Polygonizer (GCP), a method designed to address key challenges in mask-based polygonization.
First, inspired by the success of contour-based approaches, we incorporate a transformer-based contour regression module to refine the initial contours extracted from binary masks.
Second, to mitigate the performance degradation commonly observed in traditional polygon simplification of mask contours, we propose a collinearity-aware polygon simplification method that effectively simplifies initial mask contours. Specifically, this method optimizes a combinatorial objective that jointly minimizes the total distance between sampled contour points and the resulting polygon while reducing the number of selected edges. The optimization is performed using Dynamic Programming \cite{bellman1966dynamic}, ensuring a globally optimal solution.
Furthermore, the minimized objective can be seamlessly incorporated into the neural network as a loss function to supervise the upstream contour regression module, creating a more cohesive and efficient polygon generation pipeline.

The main contributions of this paper can be summarized as follows:
\begin{itemize}

    \item We propose GCP, a novel mask-based polygonal building mapping framework. It integrates a polyline regression module to refine contours extracted from building masks and utilizes a GCP simplification module to effectively capture piecewise linear building geometries.
    
    \item We evaluate the proposed GCP on two widely used public benchmarks, and experimental results show that it achieves state-of-the-art performance.
    
    \item The proposed GCP simplification algorithm is evaluated as a general-purpose polyline simplification method for building polygonization. Results show that it outperforms classical techniques like the Douglas-Peucker algorithm, highlighting its potential for broader applications.
    
\end{itemize}

%% file: content/related_works.tex
This section presents an overview of prior research relevant to our proposed method. We begin by examining a variety of approaches to polygonal building mapping. Subsequently, we discuss building mask extraction techniques, analyzing the advantages and disadvantages of two distinct pipelines: semantic segmentation and instance segmentation.

\subsection{Polygonal Building Mapping}
The purpose of polygon building mapping is to extract the building footprint from remote sensing images in polygon format. The extracted polygons should accurately fit the real-world building contours while being sufficiently simple, with a minimal number of edges. Polygonal building mapping is closely related to the task of building segmentation, which extracts building footprints as binary masks. Both methods aim to extract building content but present the results in vector and raster formats, respectively.

\subsubsection{Mask-based Methods}
Building on the concept of converting building masks into building polygons, previous studies seek to establish a connection between these two formats.
Notable studies include a method outlined in \cite{zhao2018building}, which uses a Mask R-CNN \cite{he2017mask} model to first extract binary building masks. These masks are then converted into simplified polygons using the Douglas-Peucker algorithm, followed by optimization with the Minimum Description Length (MDL) principle.
Wei et al. \cite{wei2019toward} propose a series of empirical, hand-crafted polygon refinement processes, such as removing overly sharp angles and merging parallel lines, to enhance the simplified polygon results generated by the Douglas-Peucker algorithm.
ASIP \cite{li2020approximating} proposes a method to partition a probability outputs from a semantic segmentation model into polygons.
It first generates and initial polygon partition of the masks using KIPPI \cite{bauchet2018kippi}, and then leverages either a merging or a splitting operator to greedily minimize a predefined energy function which considers the fidelity and complexity of each of the polygon facet.
FFL \cite{girard2021polygonal} begins by predicting an auxiliary frame field map to capture the orientations of each pixel relative to the building walls. The subsequent polygonization process encompasses several postprocessing steps, including contour extraction via Marching Square algorithm, contour refinement using an Active Contour Method (ACM) \cite{kass1988snakes}-like approach, corner-aware polyline simplification based on Douglas-Peucker, and more.

In general, the primary focus of these mask-based methods is on converting building masks into their polygonal format. This conversion often employs empirical, local, greedy, or heuristic algorithms to simplify the contours of the building masks. 
Additionally, these methods cannot be integrated into neural networks as a trainable module to be optimized simultaneously with semantic and instance segmentation models, which limits their effectiveness.

A core focus of the proposed GCP is to address these challenges by developing a globally optimal polygon simplification module. This module is not only robust but also seamlessly integrates into upstream networks, enabling a more cohesive polygon generation pipeline.

\subsubsection{Contour-based Methods}
Unlike mask-based methods that generate initial polygon contours from mask predictions, contour-based methods \cite{hu2023polybuilding,zhang2024hit,wei2023buildmapper} directly predict ordered building contours following an object detection pipeline.
One key advantage of this approach over mask-based methods is its ability to eliminate the zigzag patterns caused by the discreteness of binary masks. However, similar to mask-based methods, a major challenge remains: simplifying the densely sampled predicted contours around building corners.
A common solution presented in PolyBuilding \cite{hu2023polybuilding} enhances point predictions by adding a classification head to each and using a threshold to filter out redundant points that do not correspond to building vertices. However, this solution treats each point independently, lacking awareness of the overall building geometry, which may result in missing or redundant vertices.

\subsubsection{Connection-based Methods}
Connection-based methods \cite{zorzi2022polyworld,zorzi2023re,wei2024lines} begin by predicting the position of a set of primitives such as vertices or edges of the buildings, and then learn to order them to form polygons according to their pair-wise connectivity.
PolyWorld \cite{zorzi2022polyworld} first predicts the candidate building vertices using a semantic segmentation networks, followed by a threshold-based filtering and a non-maximum supression (NMS) \cite{hosang2017learning} operations.
It then learns to predict a permutation matrix that capture the pair-wise connection strength among these candidate vertices, and finally generate the polygons by solving a linear assignment sum problem.
Line2Poly  \cite{wei2024lines} tries to improve over this line of method by taking the edges of building, instead of vertices, as the primitive, so as to reduce the angle-wise inaccuracy caused by individual vertices prediction.
To form the final polygon, Line2Poly predicts a edge-wise adjacent matrix to capture the connection strength between the predicted edges.

\subsubsection{Sequential Modeling-based Methods}
Sequential modeling-based methods are similar to connection-based methods in terms of the identification of initial primitives.
But instead of predicting the connectivity of these primitives,
they directly decode the final polygon, and output the final primitives step-by-step, using a sequence prediction model such as Recurrent Neural Networks (RNN) \cite{medsker2001recurrent} and Transformer \cite{vaswani2017attention}.
For example, PolyMapper \cite{li2019topological} first generates a vertices mask for each of the building instance, and then order these vertices using a RNN.
OEC-RNN \cite{huang2021oec} proposes to extract an edge attention map in addition to the vertices mask to guide the squential decoding process conducted by ConvLSTM \cite{shi2015convolutional}.
P2PFormer \cite{zhang2024p2pformer} introduces the utilization of various primitives, including edges and corner triplets, in addition to vertices. It incorporates a transformer as the sequential decoder and introduces a group query mechanism. This mechanism allows the network to concentrate on different regions associated with primitives that may have multiple endpoints.

\subsection{Building Mask Extraction}
The extraction of binary building masks generally employs two distinct frameworks. The first framework utilizes semantic segmentation models to distinguish between building and non-building pixels.
In scenarios that require the identification of individual building instances, it becomes essential to recognize all connected components, treating each as a separate entity.
However, this approach may lead to the merging of closely situated buildings into a single structure, especially when dealing with images of lower resolution.

To address these issues, some researchers have turned to instance segmentation models, which generate separate binary masks for each building instance. A prominent approach in building extraction is Mask R-CNN \cite{he2017mask}, which extends Faster R-CNN \cite{ren2016faster} by incorporating a mask prediction head for binary segmentation within each detected bounding box. Following the introduction of Transformer \cite{vaswani2017attention} and DETR-based object detection frameworks \cite{carion2020end}, transformer-based instance segmentation models such as MaskFormer \cite{cheng2021per}, Mask2Former \cite{cheng2022masked}, and Mask-DINO \cite{li2022mask} have emerged. These models provide simplified architectures and enhanced performance.
Considering these advancements, our GCP method is built on the Mask2Former framework.

%% file: content/methods.tex
\begin{figure*}[htp]
    \centering
    \includegraphics[width=1.0\linewidth]{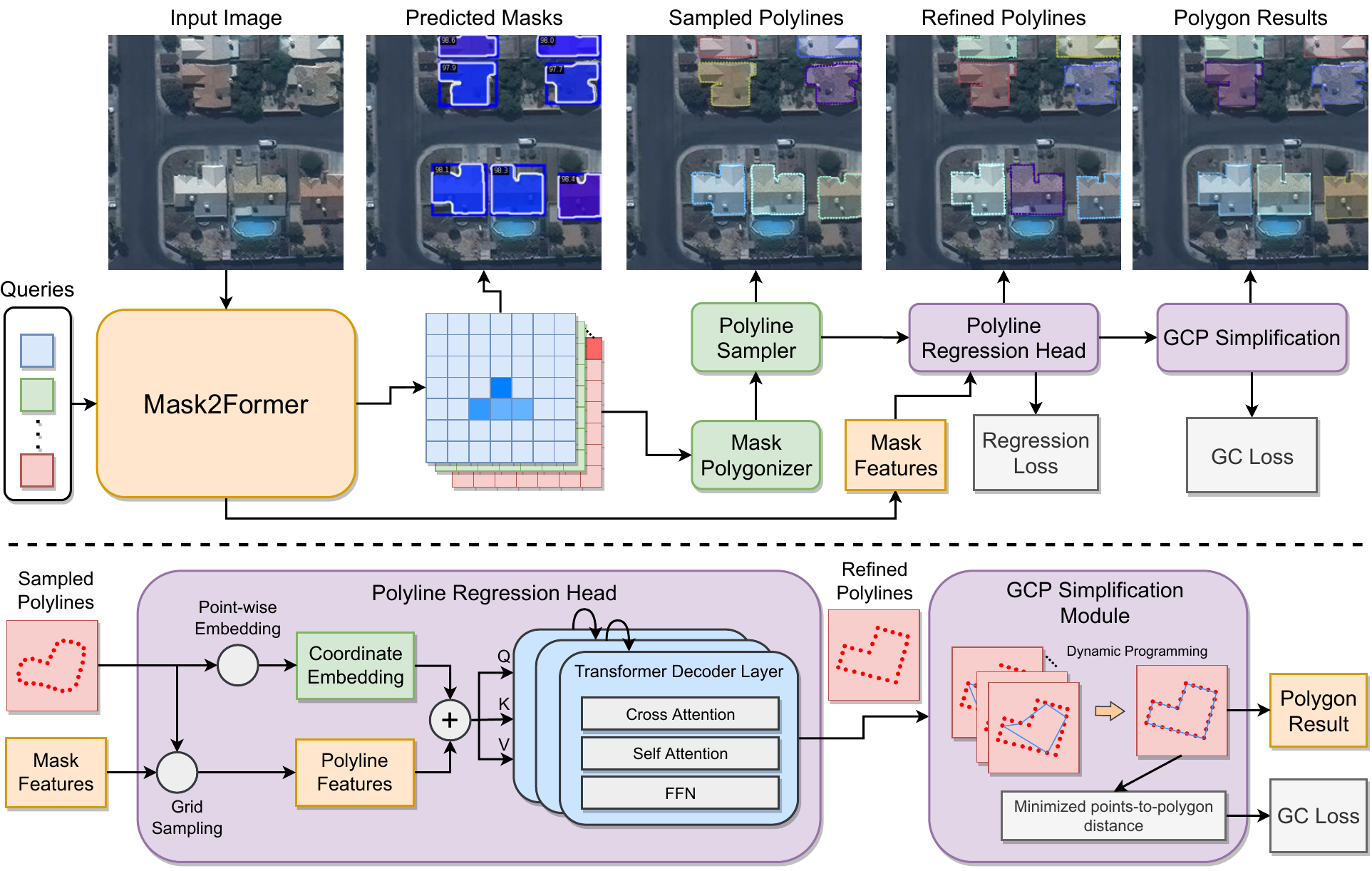}
    \caption{
        Flowchart of the proposed GCP. The upper section illustrates the overall pipeline, while the bottom section provides a detailed look at the architecture of two specific modules.
        }
    \label{fig:pipeline}
\end{figure*}

The workflow of the proposed GCP is illustrated in Figure \ref{fig:pipeline}.
The entire pipeline consists of three main stages: initial contour generation, polyline regression, and global collinearity-aware polygon simplification. These components are detailed in Sections \ref{sec:init_contour} to \ref{sec:GCPS}.
Following this, we describe the loss functions used to train GCP in Section \ref{sec:loss} and provide key implementation details in Section \ref{sec:implementation}.

\begin{figure}[htp]
    \centering
    \includegraphics[width=1.0\linewidth]{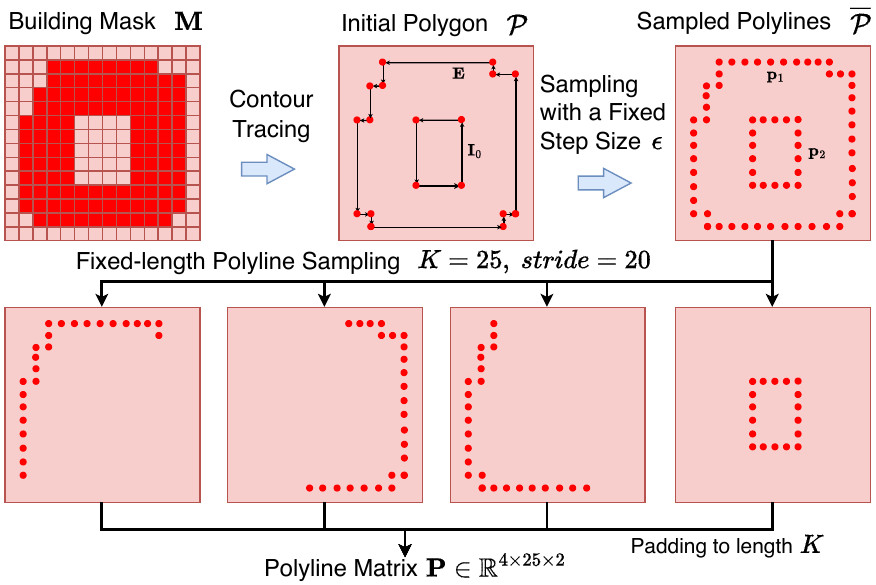}
    \caption{
        Illustration of the initial contour generation process. 
        }
    \label{fig:initial_contour}
\end{figure}

\subsection{Initial Contour Generation}
\label{sec:init_contour}
The whole pipeline begins by the generation of the initial contour corresponds to the binary masks.
Fig. \ref{fig:initial_contour} illustrates this process.
Given an input image, the utilized Mask2Former model will generate $N$ binary masks $\{\mathbf{M}_i\}_{i=1}^{N}$, each of which potentially corresponds to a single building instance in the image.
Here $N$ is a pre-defined number of queries.
Each of the predicted masks $\mathbf{M} \in \mathbb{R}^{H\times W}$ is converted to a polygon representation $\mathcal{P} = \{\mathbf{E}, \{\mathbf{I}_{0}, \mathbf{I}_{1},...\}\}$ using the contour tracing \cite{suzuki1985topological} algorithm, where $\mathbf{E} \in \mathbb{R}^{\lvert \mathbf{E} \rvert \times 2}$ denotes the coordinates of the exterior polyline of $\mathcal{P}$, organized in anti-clockwise order, and $\lvert \mathbf{E} \rvert$ is the number of points in $\mathbf{E}$.
$\mathbf{E}$ is a closed polyline, ensuring $\mathbf{E}_{0} = \mathbf{E}_{\lvert \mathbf{E} \rvert}$.
$\{\mathbf{I}_{0}, \mathbf{I}_{1},...\}$ denote the interior polyines or ``holes'' inside $\mathbf{E}$, also closed like $\mathbf{E}$.
Note that if multiple connected components exist in $\mathbf{M}$, the component with the largest area is selected to derive the polygon representation $\mathcal{P}$.

Subsequently, we uniformly sample points along each of the polyline within $\mathcal{P}$ using a fixed step size $\epsilon$ for both exterior and interior polylines, and generate a set of polylines $\overline{\mathcal{P}} = \{\overline{\mathbf{p}}_1, \overline{\mathbf{p}}_2,...\overline{\mathbf{p}}_{\lvert \overline{\mathcal{P}} \rvert}\}$.
To prevent sampled polylines from exceeding the maximum length, we set an upper limit of $L_{\max} = 512$ for each polyline. If a polyline contains more than $L_{\max}$ points, we uniformly downsample it to exactly $L_{\max}$ points to obtain $\overline{\mathbf{p}}_i$.
Since each sampled polyline $\overline{\mathbf{p}}_{i}$ can vary in length, making them unsuitable for processing by neural networks,
we apply sliding windows of size $K$ to segment each polyline in $\overline{\mathcal{P}}$.
This results in a polyline matrix
$\mathbf{P} \in \mathbb{R}^{M \times K \times 2}$,
where $M$ represents the total number of fixed-length polylines sampled across $\overline{\mathcal{P}}$.
Polylines shorter than $K$ are padded to match this size.
Each polyline $\mathbf{P}_{i} \in \mathbb{R}^{K \times 2}$ has a fixed length, fixed distance between two adjacent points, and can be either closed or open.
In the latter case, a longer polyline from \(\overline{\mathcal{P}}\) is divided into multiple segments by the sliding window, representing different segments of the original polyline.

\subsection{Polyline Regression}
Owing to the above described process, each binary mask of a building instance is represented by a polyline matrix $\mathbf{P}$ that feature the contour of the mask.
Such a representation is usually coarse and cannot well capture the intricate details around the building corners.
To address this problem,
a simple transformer decoder is used as a polyline regression head to refine the sampled polylines in $\mathbf{P}$, as illustrated in the bottom left part of Fig. \ref{fig:pipeline}.
The transformer decoder comprises three layers, each mirroring the decoder configuration in Mask2Former, with both cross-attention and self-attention layers.
To this end, the polyline regression head takes $\mathbf{P}$ as input, and output a refined polyline matrix $\tilde{\mathbf{P}}$ according to the following equation:
\begin{equation}
\label{eq:reg_head}
\tilde{\mathbf{P}} = RegHead(GridS(\mathbf{P}, \mathbf{F}) + Embed(\mathbf{P})) + \mathbf{P}.   
\end{equation}
Here $\mathbf{F} \in \mathbb{R}^{C\times H\times W}$ denotes the features map extracted by Mask2former.
$GridS(\cdot, \cdot)$ denotes the grid sampling technique, which samples the features over $\mathbf{F}$ corresponding to the coordinates of $\mathbf{P}$ using billinear interpolation.
$Embed(\cdot)$ maps any 2-dimensional coordinates to a high-dimensional feature vector.
$RegHead(\cdot)$ denotes the transformer decoder that maps the extracted features to an offset value that refines the original polylines $\mathbf{P}$.

During the inference phase, the refined polyline matrix $\tilde{\mathbf{P}}$ is reconstructed into a closed polygon by reversing the fixed-length polyline sampling process illustrated in Fig. \ref{fig:initial_contour}. Repetitively sampled points are merged by averaging their positions. Each reassembled polygon then undergoes a polygon simplification process introduced in Section \ref{sec:GCPS} to generate the final polygon predictions.

\subsection{Global Collinearity-aware Polygon Simplification}
\label{sec:GCPS}
Ideally, the refined polyine matrix $\tilde{\mathbf{P}}$ gives us an accurate representation of the building contours.
However, these refined polylines often contain a high density of redundant points.
To achieve a more simplified polygon representation, we develop a global collinearity-aware polygon simplification module.

\subsubsection{Problem Formulation}

Similar to Douglas-Peucker algorithm, the proposed polyline simplification module take as input an arbitrary polyline:
$$\mathbf{p}=(p_1, p_2,..., p_T) = ((x_1, y_1), (x_2, y_2),... (x_T, y_T)).$$
Here $T \geq 2$, is the length of the polyline, where $p_i = (x_i, y_i)$  denotes the coordinates of each point.
$\mathbf{p}$ can be either closed ($p_1 = p_T$) or open ($p_1 \neq p_T$).

The output of the module is a \emph{simplified polyline} $\hat{\mathbf{p}}$, which is a sub-sequence of $\mathbf{p}$:
\begin{equation}
\label{eq:definition}
\hat{\mathbf{p}} = (p_{t_1}, p_{t_{2}},..., p_{t_{m}}),
\end{equation}
where $(t_1, t_2,..., t_m)$, satisfying $1 = t_1 < t_2 < ..., t_m = T$, represents a list of indices that identify the sub-sequence.
$m \geq 2$ is the number of the chosen indexes.
Note that the first and the last points of $\mathbf{p}$ are always included.
Our goal is to find an optimal simplified polyline $\hat{\mathbf{p}}^{*}$ that capture the intrinsic geometry of $\mathbf{p}$.

\subsubsection{Collinearity-aware Objective}
Our core insight in searching for the optimal $\hat{\mathbf{p}}^{*}$ is to maximally preserve the collinearity of consecutive point sequences within $\mathbf{p}$.
We define $\mathcal{L}(\mathbf{p}, \hat{\mathbf{p}})$ as a measurement of distance:
\begin{equation}
    \label{eq:dis}
    \mathcal{L}(\mathbf{p}, \hat{\mathbf{p}}) = \sum_{i=1}^{m-1} \sum_{j=t_{i} + 1}^{t_{i+1} - 1} \
    dis(p_{j}, \overline{p_{t_i}p_{t_{i+1}}}).
\end{equation}
Here $dis(p_{j}, \overline{p_{t_i}p_{t_{i+1}}})$ denotes the distance from point $p_j$ to line $\overline{p_{t_i}p_{t_{i+1}}}$.
Intuitively, $\mathcal{L}(\mathbf{p}, \hat{\mathbf{p}})$ capture the sum of distance from each point of $\mathbf{p}$ to the segment that covers it according to $\hat{\mathbf{p}}$.

A small $\mathcal{L}(\mathbf{p}, \hat{\mathbf{p}})$ indicates that most of the consecutive points in $\mathbf{p}$ that lie in the same line are well represented by the corresponding segments that are chosen in $\hat{\mathbf{p}}$,
which means the inherent collinearity among points are well preserved.
To this end, we define our optimization objective for searching the optimal $\hat{\mathbf{p}}^{*}$ as:
\begin{equation}
    \hat{\mathbf{p}}^{*} = \mathop{argmin}_{\hat{\mathbf{p}}} \ \mathcal{L}(\mathbf{p}, \hat{\mathbf{p}}).
\end{equation}
However, this definition is not well-posed, as it allows for a trivial solution: the original polyline $\mathbf{p}$ itself, since $\mathcal{L}(\mathbf{p}, \mathbf{p}) = 0$.
This contradicts our intent that a ``simplified polyline'' should be minimalistic, ideally comprising just a few segments. To address this, we introduce a regularization term designed to penalize the inclusion of additional points in $\hat{\mathbf{p}}$:
\begin{equation}
    \label{eq:objective}
    \hat{\mathbf{p}}^{*} = \mathop{argmin}_{\hat{\mathbf{p}}} \ \mathcal{L}(\mathbf{p}, \hat{\mathbf{p}}) + \lambda \lvert \hat{\mathbf{p}} \rvert,
\end{equation}
where $\lvert \hat{\mathbf{p}} \rvert = m$ is the number of the selected points in $\hat{\mathbf{p}}$, and $\lambda$ is a coefficient used to balance the two terms.
By minimizing this objective function, 
we ensure that the simplified polygon $\hat{\mathbf{p}}^{*}$ remains compact, yet effectively captures the underlying piecewise linear geometry of $\mathbf{p}$.

\subsubsection{Optimization by Dynamic Programming}
\begin{figure*}[htp]
    \centering
    \includegraphics[width=0.95\linewidth]{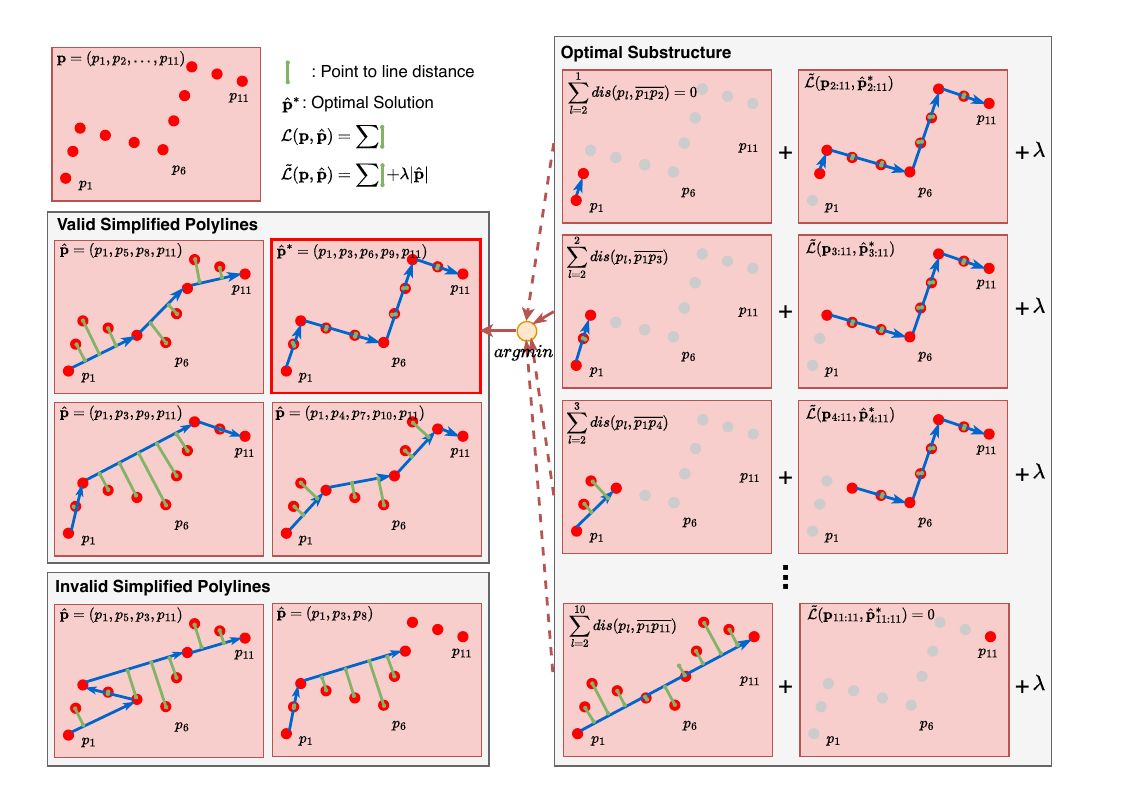}
    \caption{
    Illustration of the proposed global collinearity-aware polyline simplification method.
    For a given polyline to be simplified $\mathbf{p}$,
    The middle left part of the figure lists $4$ different possible simplification plans.
    The bottom left part shows $2$ invalid simplification plans according to the definition in Eq. (\ref{eq:definition}).
    The right part of the figure demonstrates how the optimal solution is achieved by combining the solutions to its sub-problems according to Eq. (\ref{eq:dp}). Such a property is termed ``optimal substructure'' in Dynamic Programming.
    }
    \label{fig:dynamic_programming}
\end{figure*}

Eq. \ref{eq:objective} is a combinatorial optimization problem defined in discrete space.
Here we provide an optimal solution to it using dynamic programming technique \cite{bellman1966dynamic}.
For clarity, we first define:
\begin{equation}
    \tilde{\mathcal{L}}(\mathbf{p}, \hat{\mathbf{p}}) = \mathcal{L}(\mathbf{p}, \hat{\mathbf{p}}) + \lambda \lvert \hat{\mathbf{p}} \rvert.
\end{equation}
Given two integers $i, k$, satisfying $1 \leq i < T$ and $1 < k \leq T-i$,
it can be observed that the bellow equation holds:
\begin{align}
\label{eq:dp}
\begin{split}
    \tilde{\mathcal{L}}(\mathbf{p}_{i:i+k}, \hat{\mathbf{p}}_{i:i+k}^{*}) = & \mathop{min}_{i < j \leq i+k} \
    \tilde{\mathcal{L}}(\mathbf{p}_{j:i+k}, \hat{\mathbf{p}}_{j:i+k}) \\
    & \quad + \sum_{l=i+1}^{j-1} dis(p_{l}, \overrightarrow{p_{i} p_{j}}) + \lambda.
\end{split}
\end{align}
Here $\mathbf{p}_{i:i+k} = (p_{i}, p_{i+1},..., p_{i+k})$ denotes a consecutive sub-sequence of $\mathbf{p}$, $\hat{\mathbf{p}}_{i:i+k}$ is the corresponding simplified polyline,
and $\hat{\mathbf{p}}_{i:i+k}^{*}$ defines the specific simplification plan that yields the minimum $\tilde{\mathcal{L}}$ value.
$\tilde{\mathcal{L}}(\mathbf{p}_{i:i+k}, \hat{\mathbf{p}}_{i:i+k}^{*})$ defines a sub-problem of the origin polyline simplification problem, modifying the input polyline line $\mathbf{p}$ to $\mathbf{p}_{i:i+k}$.
Specifically, to ensure Eq. \ref{eq:dp} remains meaningful when $j = i + k$, we define:
\begin{equation}
    \tilde{\mathcal{L}}(\mathbf{p}_{i:i}, \hat{\mathbf{p}}_{i:i}^{*}) = 0.
\end{equation}
Consequently, when $k=1$, the solution is obvious:
\begin{equation}
    \label{eq:init}
    \tilde{\mathcal{L}}(\mathbf{p}_{i:i+1}, \hat{\mathbf{p}}_{i:i+1}^{*}) = \lambda.
\end{equation}
To this end, $\tilde{\mathcal{L}}(\mathbf{p}_{i:i+2}, \hat{\mathbf{p}}_{i:i+2})$ can be resolved by combining Eq. (\ref{eq:init}) and Eq. (\ref{eq:dp}).
By repeating this process, we are able to solve the sub-problem $\tilde{\mathcal{L}}(\mathbf{p}_{1:T}, \hat{\mathbf{p}}_{1:T})$,
which is equivalent to the original problem $\tilde{\mathcal{L}}(\mathbf{p}, \hat{\mathbf{p}})$.
The optimization process is illustrated in Fig. \ref{fig:dynamic_programming}.

\subsection{Loss Functions}
\label{sec:loss}
This section presents the loss functions for supervising the training of the proposed GCP.

\subsubsection{Regression Loss}
During the training of Mask2Former, each predicted binary mask $\mathbf{M}$ is assigned with a ground truth polygon $\overline{\mathbf{p}}$ via Hungarian algorithm \cite{kuhn1955hungarian}.
This ground truth polygon can be used to supervise the predicted polylines $\tilde{\mathbf{P}}$ generated by Eq. (\ref{eq:reg_head}).
Since $\tilde{\mathbf{P}} \in \mathbb{R}^{M \times K \times 2}$ may contains multiple polylines, we randomly sample a single one $\tilde{\mathbf{p}}$ for loss calculation.
Then, we calculate a Hungarian assignment between $\mathbf{p}$ and $\overline{\mathbf{p}}$ in a point-wise manner.
The $\mathbf{p}$ here refers to the original polyline corresponds to $\tilde{\mathbf{p}}$ without the refinement process in Eq. (\ref{eq:reg_head}).
Only the vertex pairs with distances less than a predefined threshold $\epsilon=15$ are selected. For these matched vertex pairs, $\tilde{\mathbf{p}}^{matched}$ and $\overline{\mathbf{p}}^{matched}$, we compute a smooth L1 loss between them.
\begin{equation}
    \mathcal{L}_{vert} = SmoothL1(\tilde{\mathbf{p}}^{matched}, \overline{\mathbf{p}}^{matched}).
\end{equation}
Besides, we calculate an angular loss similar to the method used in PolyWorld \cite{zorzi2022polyworld}.
For each set of three consecutive matched points, we calculate the angular difference and then adopt the maximum difference per polyline to determine the angular loss:
\begin{equation}
    \mathcal{L}_{ang} = \mathop{max}_{(u,v,w)} \lvert \tilde{\angle}_{u,v,w} - \overline{\angle}_{u,v,w}\rvert,
\end{equation}
where $\tilde{\angle}_{u,v,w}$ and $\overline{\angle}_{u,v,w})$ indicate the angles formed by three consecutive points indexed by $u$, $v$ and $w$ of $\tilde{\mathbf{p}}^{matched}$ and $\overline{\mathbf{p}}^{matched}$, respectively.

\subsubsection{Global Collinearity Loss}
To make the entire polyline refinement and simplification pipeline more cohesive, we incorporate supervision to the refined polylines conditioned on the final decoded polygon results.
Additionally, we aim to enhance the collinearity of the predicted polygons. 
To achieve this, we calculate a global collinearity loss for each decoded polygon.
Specifically, given a refined polyline $\tilde{\mathbf{p}}$ sampled from $\tilde{\mathbf{P}}$, and its simplified result $\tilde{\mathbf{p}}^{*}$ obtained using Eq. (\ref{eq:objective}), we define the global collinearity loss $\mathcal{L}_{gc}$ as follows:
\begin{equation}
    \mathcal{L}_{gc} = 
    \mathcal{L}(\tilde{\mathbf{p}}, \tilde{\mathbf{p}}^{*}),
\end{equation}
which is our unnormalized objective function value conditioned on the optimal solution $\tilde{\mathbf{p}}^{*}$.
According to Eq. \ref{eq:dis}, it can be reformulated as:
\begin{equation}
    \mathcal{L}_{gc} = \sum_{i=1}^{\lvert \tilde{\mathbf{p}}^{*} \rvert-1} \sum_{j=t_{i}^{*} + 1}^{t_{i+1}^{*} - 1} \
    dis(\tilde{p}_{j}, \overline{\tilde{p}_{t_i^{*}}\tilde{p}_{t_{i+1}^{*}}}),
\end{equation}
where $\{t_i^{*}\}$ denotes selected indexes of $\tilde{\mathbf{p}}$ corresponds to $\tilde{\mathbf{p}}^{*}$.
Since $\mathcal{L}_{gc}$ is essentially a sum of point-to-line distance, it remains differentiable with respect to $\tilde{\mathbf{p}}$.

\subsection{Implementation of GCP}
\label{sec:implementation}
This subsection provides key implementation details of the proposed GCP simplification algorithm. Its pseudo-code is presented in Algorithm \ref{alg:gcp}. Section \ref{sec:computation_complexity} analyzes its computational complexity, while Section \ref{sec:GCP_acceleration} discusses various acceleration strategies.

\subsubsection{Computational Complexity}
\label{sec:computation_complexity}
To solve Eq. (\ref{eq:objective}), we first precompute the point-to-line distances for all point-line pairs (line 17 in Algorithm \ref{alg:gcp}). We then progressively solve Eq. (\ref{eq:dp}) from \( k = 2 \) to \( T - 1 \) (lines 18--31 in Algorithm \ref{alg:gcp}).  

To improve computational efficiency, we introduce a maximum step size \( k_{\max} \leq T \), which constrains the range of \( j \) in Eq. (\ref{eq:dp}), ensuring \( j - i \leq k_{\max} \). Intuitively, this means that the longest edge of the simplified polygon \( \hat{\mathbf{p}} \) can span at most \( k_{\max} \) consecutive points.  
This restriction reduces the time complexity of solving Eq. (\ref{eq:objective}) from \( O(T^3) \) to \( O(T \cdot k_{\max}^{2} + T^2 \cdot k_{\max}) \), where
\( O(T \cdot k_{\max}^{2}) \) accounts for precomputing the point-to-line distances, and \( O(T^2 \cdot k_{\max}) \) corresponds to the iterative solution of Eq. (\ref{eq:dp}).

\begin{algorithm}
\caption{GCP Simplification Algorithm}
\label{alg:gcp}
\begin{algorithmic}[1]  

\Require A polyline $\mathbf{p} \in \mathbb{R}^{T \times 2}$, maximun step size $k_{max}$, regularization coefficient $\lambda$ in Eq. \ref{eq:objective}.
\Ensure Simplified polyline $\mathbf{p}^{*}$.

\Function{Point2Line}{$\mathbf{p}_1$, $\mathbf{p}_2$, $\mathbf{p}_3$}
    \State Set $d$ to the distance from point $\mathbf{p}_3$ to line $\overline{\mathbf{p}_1 \mathbf{p}_2}$.
    \State \Return $d$
\EndFunction

\Function{CalDisMat}{$\mathbf{p}$, $k_{max}$}
    \State // Get point-to-vector distance in Eq. 
    \ref{eq:dis}.
    \State Initialize $\mathbf{D}$ as a zero matrix of shape $(T, T)$.
    \For{$i=1$ to $T$}
        \For{$j=i+1$ to $min(T, i+k_{max})$}
            \For{$k=i+1$ to $j-1$}
                \State $\mathbf{D}_{i,j} \gets \mathbf{D}_{i,j} + \text{Point2Line}(\mathbf{p}_i, \mathbf{p}_j$, $\mathbf{p}_k)$.
            \EndFor
        \EndFor
    \EndFor
    \State \Return $\mathbf{D}$
\EndFunction
\State $\mathbf{D} \gets \text{CalDisMat}(\mathbf{p}, k_{max})$.
\State Define $\mathbf{\Lambda} \in \mathbb{R}^{T \times T}$ with all elements equal to $\lambda$.
\State $\mathbf{C} \gets \mathbf{D} + \mathbf{\Lambda}$. // Cost Matrix.
\State Initialize $\tilde{\mathbf{L}},\mathbf{F} \in \mathbb{R}^{T \times T}$ to zero matrices.
\For{$k=1$ to $T$}
    \For{$i=1$ to $T-k$}
        \State $\tilde{\mathbf{L}}_{i,i+k} \gets +\infty$.
        \For{$j=i+1$ to $i + \mathop{min}(k, k_{max})$}
            \If{$\tilde{\mathbf{L}}_{j,i+k} + \mathbf{C}_{i,j} < \tilde{\mathbf{L}}_{i,i+k}$}
                \State $\tilde{\mathbf{L}}_{i,i+k} \gets \tilde{\mathbf{L}}_{j,i+k} + \mathbf{C}_{i,j}.$ // Eq. \ref{eq:dp}.
                \State $\mathbf{F}_{i,i+k} \gets j$ // Record the index.
            \EndIf
        \EndFor
    \EndFor
\EndFor
\State Initialize $\mathbf{p}^{*}$ as an empty set. 
\State $\mathbf{p}_{1}^{*} \gets 1.$
\State Set $j \gets 1, i \gets 2$.
\While{$j < T$}
    \State $\mathbf{p}^{*}_{i} \gets \mathbf{F}_{j,T}$.
    \State $j \gets \mathbf{F}_{j,T}.$
    \State $i \gets i + 1$.
\EndWhile
\State \Return $\mathbf{p}^{*}$

\end{algorithmic}
\end{algorithm}

\subsubsection{Acceleration of GCP Simplification}
\label{sec:GCP_acceleration}
Algorithm \ref{alg:gcp} presents a naive implementation of the proposed GCP simplification algorithm, which extensively utilizes loop structures to compute pairwise point-to-line distances and iteratively solve Eq. \ref{eq:dp}.
However, it is important to note that most of these loops can be parallelized and vectorized, as each iteration is independent of the previous ones. Specifically, the only necessary loop structures that cannot be parallelized are the loop in line 21, which iterates over the sub-sequence length of $\mathbf{p}$ in Eq. \ref{eq:dp}, and the loop in line 35, which sequentially decodes the simplified polyline.

Thus, Algorithm \ref{alg:gcp} can be significantly accelerated by parallelizing and vectorizing the computation of key variables, including $\mathbf{D}$, $\tilde{\mathbf{L}}$, and $\mathbf{F}$. Additionally, the GCP simplification method can be further optimized through batch processing and GPU acceleration.
For additional implementation details, please refer to our published code.

%% file: content/experiments.tex
In this section, we present and analyze the experimental results to validate the effectiveness of the proposed GCP framework.

\subsection{Datasets and Evaluation Metrics}
\subsubsection{Datasets}

Two widely used public benchmarks were employed in our experiments: the CrowdAI Mapping Challenge \cite{crowdai2023mapping} and WHU-Mix (vector) \cite{wei2023buildmapper} datasets.

The CrowdAI dataset consists of aerial RGB images with a ground sampling distance of approximately $0.2$m. It includes a training split with $280,741$ images and a validation split with $60,317$ images, all sized $300 \times 300$ pixels.
We chose this dataset to evaluate the proposed method because it is one of the most commonly used in the field of polygonal building mapping, providing numerous baseline comparisons.
However, it is important to note a valid study by \cite{adimoolam2023efficient} that highlights a potential data leakage issue in the CrowdAI dataset, which may reduce the reliability of comparative results.

The WHU-Mix (vector) dataset comprises a large collection of aerial and satellite images from various sources, including WHU \cite{ji2018fully}, CrowdAI, Open AI \cite{codalab20100}, SpaceNet \cite{van2018spacenet}, and Inria \cite{maggiori2017can}. The dataset covers five continents and spans over 10 regions worldwide, with a spatial resolution ranging from $0.09$m to $2.5$m. It includes a training set of $43,778$ images, a validation set of $2,922$ images, and two test sets containing $11,675$ and $6,011$ images, respectively.
Notably, the second test set does not overlap geographically with the training set, presenting a greater challenge for the generalizability of building mapping models.

\subsubsection{Evaluation Metrics}
We employed a variety of metrics to evaluate the effectiveness of our proposed method, including MS-COCO metrics, Intersection over Union (IoU), Maximum Tangent Angle Error (MTA) \cite{girard2021polygonal}, Corrected-IoU (C-IoU) \cite{zorzi2022polyworld}, and the N-ratio~\cite{zorzi2022polyworld}.

The MS-COCO metrics provide Average Precision (AP) and Average Recall (AR) values for the predicted masks across IoU thresholds ranging from 0.5 to 0.95, in increments of 0.05. Specifically, we report AP and AR at IoU thresholds of 0.5 and 0.75, denoted as AP${50}$, AP${75}$, AR${50}$, and AR${75}$.

The MTA metric measures angle-wise discrepancies between predicted and actual ground truth contours. It considers contour pairs with an IoU greater than 0.5, where the predicted contour is uniformly sampled at intervals of 0.1 pixels and mapped onto the ground truth. The tangent angle differences between consecutive points on the sampled and projected sequences are calculated, with the maximum angle difference across each contour pair serving as the MTA value.

C-IoU refines the IoU calculation by scaling it according to the relative difference in the number of predicted and ground truth vertices for each image.

Lastly, the N-ratio measures the ratio of predicted vertices to ground truth vertices. An N-ratio below 1 indicates fewer predicted vertices than the ground truth, while a ratio above 1 indicates an overprediction of vertices.

\subsection{Implementation Details}
GCP is built on the Mask2Former framework with ResNet-50 \cite{he2016deep} as the backbone.
For hyperparameters, the sampling step size $\epsilon$ and the sliding window size in Fig. \ref{fig:initial_contour} are set to 4 and 64, respectively. The regularization coefficient $\lambda$ in Eq. (\ref{eq:objective}) is set to 2 for the CrowdAI dataset and 4 for the WHU-Mix (vector) dataset. The maximum step size $k_{max}$, as discussed in Section \ref{sec:computation_complexity}, is set to 64.
All experiments are conducted on a single NVIDIA A40 GPU.

The training process for the proposed GCP model is divided into two stages. In the first stage, we train the baseline Mask2Former model. In the second stage, the parameters of the Mask2Former model are frozen, and training is focused solely on the polygonization process. At this point, only the polyline regression head and a convolutional mapping layer, which reprojects the mask features extracted by Mask2Former, remain as trainable components.

For the experiments on the CrowdAI dataset, the number of queries used to generate predicted masks with Mask2Former is set to $100$. All images are resized to $320 \times 320$ to ensure divisibility by $32$, preventing pooling issues within the network. Batch size is set to $16$. The predicted polygons are then resized back to the original $300 \times 300$ dimensions for metric calculation.
Data augmentation techniques include random horizontal and vertical flipping, as well as random rotations of 90, 180, and 270 degrees. The Mask2Former model is trained for 100 epochs using the AdamW optimizer \cite{loshchilov2017decoupled}, with an initial learning rate of $10^{-4}$, which is decayed to $10^{-5}$ after the 80th epoch.
In the second stage, GCP is trained for 80,000 iterations (roughly equivalent to 5 epochs), starting with an initial learning rate of $10^{-4}$, which is reduced to $10^{-5}$ after the 60,000th iteration.

For the experiments on the WHU-Mix (vector) dataset, the number of queries is set to 300, and all images are resized to $512 \times 512$ during both the training and inference stages. The batch size is set to 8. Similar to the CrowdAI setting, the predicted results are resized back to their original dimensions for evaluation.
In the first stage, Mask2Former is trained for 50 epochs with an initial learning rate of $10^{-4}$, which is reduced to $10^{-5}$ after the 40th epoch. In the second stage, GCP is trained for 12 epochs, starting with an initial learning rate of $10^{-4}$, decaying to $10^{-5}$ after the 9th epoch.

Since the proposed method follows an instance segmentation pipeline, the output from Mask2Former consists of multiple binary masks, which may overlap. To accurately compute metrics other than the MS-COCO metrics, we apply a confidence threshold of $0.5$ to filter out redundant masks. The remaining masks are then aggregated to generate a single mask for calculating IoU and C-IoU values.
Similarly, only predicted polygons with a confidence score higher than $0.5$ are considered when calculating MTA, C-IoU, and N-ratio.

\setlength\tabcolsep{0pt}
\begin{table*}
\footnotesize
\caption{
     Quantitative comparative results on CrowdAI dataset.
}
\renewcommand\arraystretch{1.2}
\newcolumntype{C}{>{\centering \arraybackslash}m{1.2cm}}
\newcolumntype{E}{>{\centering \arraybackslash}m{1.7cm}}
\newcolumntype{D}{>{\centering \arraybackslash}m{3.5cm}}
\begin{center}
    \begin{tabular}{ D | E | C C C C C C C C C C}
\toprule
        Methods & Backbone & AP$\uparrow$ & AP$_{50}$$\uparrow$  & AP$_{75}$$\uparrow$ & AR$\uparrow$ & AR$_{50}$$\uparrow$ & AR$_{75}$$\uparrow$ & IoU$\uparrow$ & MTA$\downarrow$ & C-IoU$\uparrow$ & N-ratio \\

\addlinespace[0.5ex]
\cline{1-12}
\addlinespace[0.5ex]
PolyMapper \cite{li2019topological} & VGG16 & 50.8 & 81.7 & 58.6 & 47.6 & 70.8 & 55.5 & 77.6 & - & 67.5 & - \\
FFL (ACM Poly) \cite{girard2021polygonal} & UResNet101 & 61.3 & 87.4 & 70.6 & 64.9 & 89.4 & 73.9 & 84.1 & 33.5 & 73.7 & 1.13 \\
PolyWorld \cite{zorzi2022polyworld} & R2U-Net & 63.3 & 88.6 & 70.5 & 75.4 & 93.5 & 83.1 & 91.3 & 32.9 & 88.2 & 0.93 \\
BuildingMapper \cite{wei2023buildmapper} & DLA-34 & 63.9 & 90.1 & 75.0 & - & - & - & - & - & - & - \\
HiT \cite{zhang2024hit} & ResNet50 & 64.6 & 91.9 & 78.7 & 75.5 & 93.8 & 83.5 & - & 31.7 &  88.6 & 1.00 \\
P2PFormer \cite{zhang2024p2pformer} & ResNet50 & 66.0 & 91.1 & 77.0 & - & - & - & - & - & - & - \\
HiSup \cite{xu2023hisup} & HRNetV2 & 79.4 & 92.7 & 85.3 & 81.5 & 93.1 & 86.7 & \textbf{94.3} & - & \textbf{89.6} & - \\
PolyBuilding \cite{hu2023polybuilding} & ResNet50 & 78.7 & 96.3 & 89.2 & 84.2 & 97.3 & 92.9 & 94.0 & 32.4 & 88.6 & 0.99 \\

\addlinespace[0.5ex]
\cline{1-12}
\addlinespace[0.5ex]

GCP (Ours) & ResNet50 & \textbf{81.5} & \textbf{96.9} & \textbf{91.2} & \textbf{85.4} & \textbf{97.6} & \textbf{93.6} & 93.4 & \textbf{31.6} & 84.0 & 0.83 \\

\bottomrule
\end{tabular}
\end{center}
\label{tab:crowd_ai}
\end{table*}

\begin{figure*}[htp]
    \centering
    \includegraphics[width=1.0\linewidth]{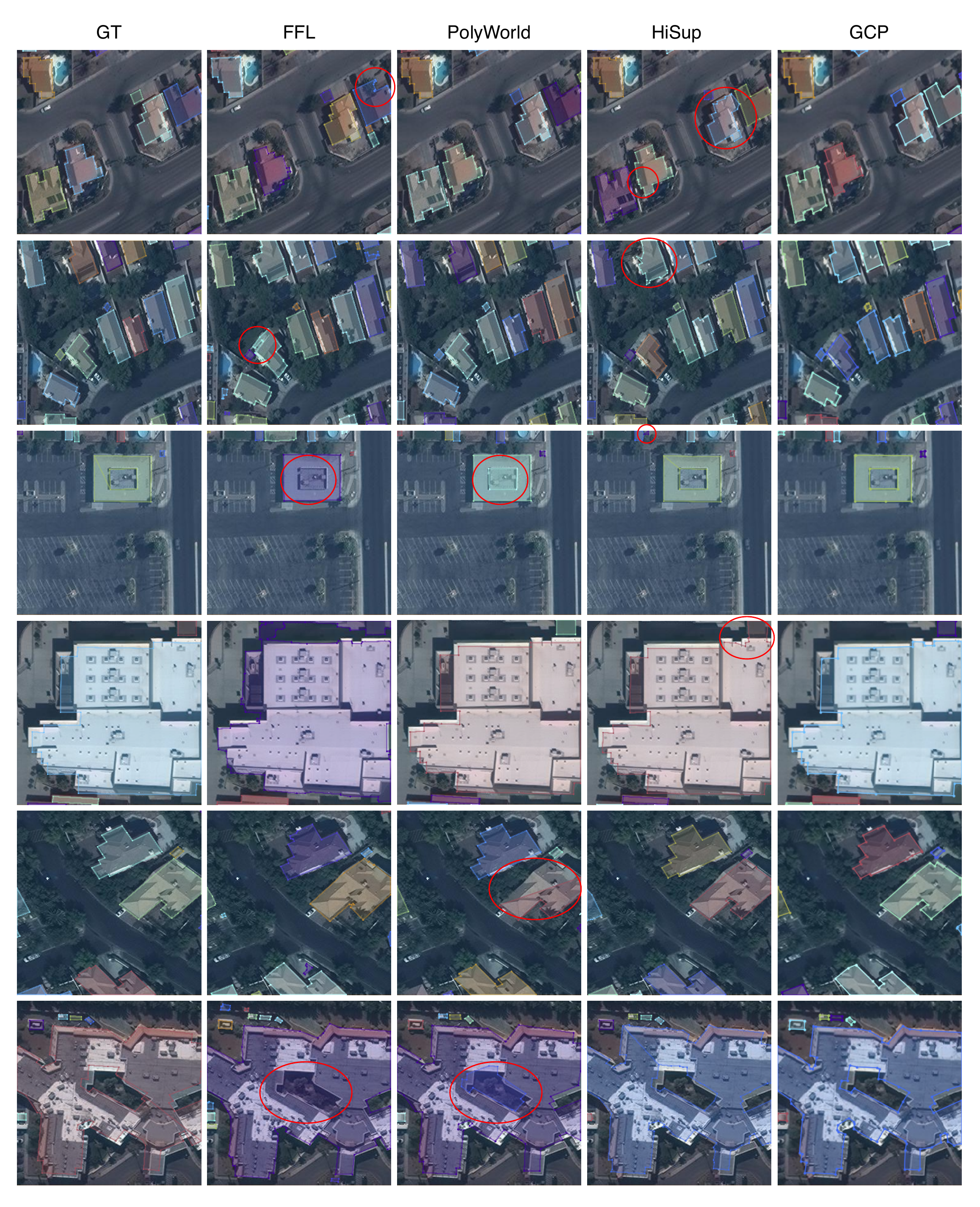}
    \caption{
        Visualized polygonal building predictions for different comparative methods on CrowdAI dataset. Red circles indicate areas where the building structure is not accurately represented.
        }
    \label{fig:crowd_ai_compare}
\end{figure*}

\subsection{Experimental Results}
In this section, we compare the proposed GCP method with the other state-of-the-art methods on CrowdAI and WHU-Mix (vector) datasets, and also conducted an ablation study to analyze the significance of each proposed network component.
The methods used for comparison include PolyMapper \cite{li2019topological}, Frame Field Learning (FFL) \cite{girard2021polygonal}, PolyWorld \cite{zorzi2022polyworld}, BuildingMapper \cite{wei2023buildmapper}, HiT \cite{zhang2024hit}, P2PFormer \cite{zhang2024p2pformer}, HiSup  \cite{xu2023hisup}, PolyBuilding \cite{hu2023polybuilding}, and Line2Poly \cite{wei2024lines}.

\subsubsection{Results on CrowdAI Datasets}
The quantitative comparative results on CrowdAI dataset is shown in Table \ref{tab:crowd_ai}.
In terms of the COCO metrics, the proposed GCP achieves AP, AP$_{50}$, and AP$_{75}$ values of $81.5$, $96.9$ and $91.2$, outperforming the previous methods.
In terms of IoU metrics, GCP outperforms most of the methods but is slightly outperformed by HiSup and PolyBuilding. This may be attributed to the aggregation of multiple masks predicted by Mask2Former into a single semantic mask, which can reduce performance.

Compared to previous methods, GCP achieves a significantly lower N-ratio, indicating that it produces a more compact polygon representation than human annotations.
This can be attributed to two key factors. First, the CrowdAI dataset annotations contain consecutive duplicate vertices, introducing unnecessary complexity and redundancy. Second, by minimizing the global collinearity loss $\mathcal{L}_{gc}$, GCP encourages the generation of polygons with fewer vertices.
As a result, for rounded buildings, which are typically annotated with densely placed vertices, GCP predicts significantly fewer vertices.

Since the predicted polygons have fewer vertices than the ground truth, GCP exhibits a lower C-IoU compared to some previous methods. However, we argue that GCP provides a compact and simplified polygon representation without compromising accuracy metrics such as AP. This property is highly desirable for many applications.

In Fig. \ref{fig:crowd_ai_compare}, we visualize the predicted polygons generated by various comparative methods alongside the proposed GCP on the CrowdAI dataset. It can be observed that different methods encounter distinct challenges.
For example, FFL and PolyWorld struggle with cases where building polygons contain internal holes, while HiSup occasionally produces polygons with sharp angles and redundant vertices. Additionally, PolyWorld may omit crucial vertices, particularly when they are located close to another building's vertex. 
In contrast, the proposed GCP method effectively addresses these issues, generating simplified yet accurate polygons.

\setlength\tabcolsep{0pt}
\begin{table*}
\footnotesize
\caption{
    Quantitative comparative results on WHU-Mix (vector) dataset.
}
\renewcommand\arraystretch{1.2}
\newcolumntype{C}{>{\centering \arraybackslash}m{1.3cm}}
\newcolumntype{D}{>{\centering \arraybackslash}m{4.0cm}}
\newcolumntype{E}{>{\centering \arraybackslash}m{1.7cm}}
\begin{center}
    \begin{tabular}{ D | E | C C C C C C C }
\toprule
        \multicolumn{9}{c}{Test1} \\
        \addlinespace[0.5ex]
        Method & Backbone & AP$\uparrow$ & AP$_{50}$$\uparrow$ & AP$_{75}$$\uparrow$ & IoU$\uparrow$ & MTA$\downarrow$ & C-IoU$\uparrow$ & N-ratio \\
        
\addlinespace[0.5ex]
\cline{1-9}
\addlinespace[0.5ex]

BuildingMapper \cite{wei2023buildmapper} & DLA-34 & 58.0 & 82.6 & 65.9 & - & - & - & - \\
Line2Poly \cite{wei2024lines} & DLA & 58.6 & 81.9 & 66.2 & - & - & - & - \\
P2PFormer \cite{zhang2024p2pformer} & ResNet50 & 60.6 & 87.3 & 68.9 & - & - & - & - \\
GCP (Ours) & ResNet50 & \textbf{62.0} & \textbf{87.6} & \textbf{69.1} & 79.6 & 32.4 & 68.8 & 0.96 \\

\addlinespace[0.5ex]
\cline{1-9}
\addlinespace[0.5ex]
\cline{1-9}
\addlinespace[0.5ex]

\multicolumn{9}{c}{Test2} \\
\addlinespace[0.5ex]
Method & Backbone & AP$\uparrow$ & AP$_{50}$$\uparrow$ & AP$_{75}$$\uparrow$ & IoU$\uparrow$ & MTA$\downarrow$ & C-IoU$\uparrow$ & N-ratio \\
\addlinespace[0.5ex]
\cline{1-9}
\addlinespace[0.5ex]

BuildingMapper \cite{wei2023buildmapper} & DLA-34 & 48.1 & 73.2 & 52.0 & - & - & - & - \\
Line2Poly \cite{wei2024lines} & DLA & 48.9 & 73.3 & 52.8 & - & - & - & - \\
P2PFormer \cite{zhang2024p2pformer} & ResNet50 & 50.7 & 79.9 & 54.4 & - & - & - & - \\
GCP (Ours) & ResNet50 & \textbf{52.9} & \textbf{80.1} & \textbf{56.5} & 77.5 & 34.2 & 65.7 & 0.94 \\
        
\bottomrule
\end{tabular}
\end{center}
\label{tab:whu_mix}
\end{table*}

\begin{figure*}[htp]
    \centering
    \includegraphics[width=1.0\linewidth]{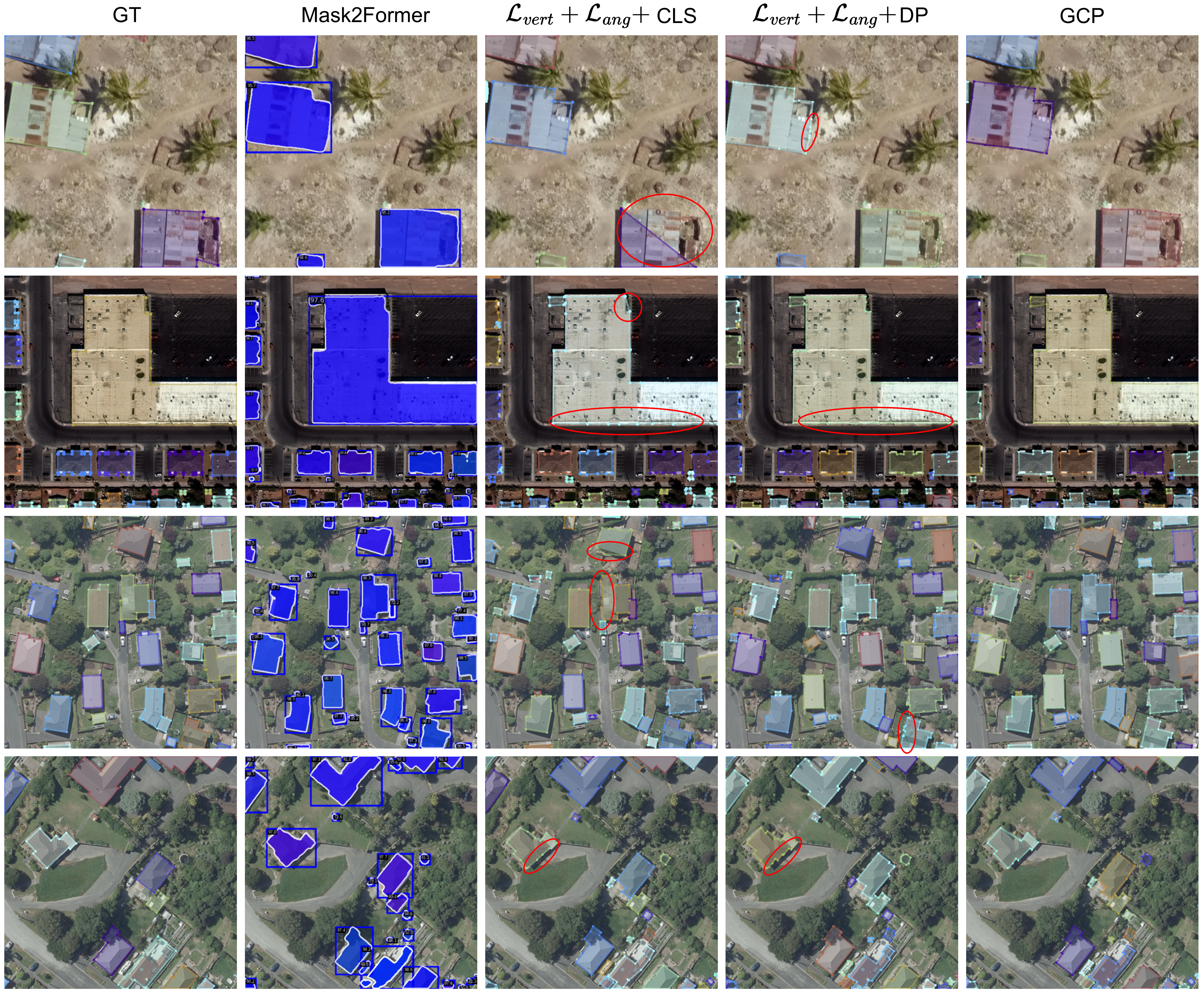}
    \caption{
Visualized polygonal building predictions for various ablated models on WHU-Mix (vector) dataset. Red circles indicate areas where the building structure is not accurately represented. The images in the first two rows pertain to the Test 1 set, whereas the last two rows feature images from the Test 2 set..
        }
    \label{fig:ablation_study}
\end{figure*}

\subsubsection{Results on WHU-Mix-Vector Datasets}

The quantitative comparative results for the two test sets of the WHU-Mix (vector) dataset are presented in Table \ref{tab:whu_mix}. In test set 1, the proposed GCP demonstrates superior performance in terms of AP, AP$_{50}$, and AP$_{75}$. The AP$_{50}$ and AP$_{75}$ values for GCP are slightly better than those of the previous state-of-the-art method, P2PFormer, with margins of 0.2 to 0.3. Notably, GCP's AP value surpasses that of P2PFormer by a more significant margin of 1.4, suggesting that the proposed method achieves higher precision at larger IoU thresholds and is potentially better at capturing detailed building geometries.

On the more challenging test set 2, the proposed GCP outperforms the second-best method by a larger margin. Specifically, GCP's AP value is higher than that of P2PFormer by 2.2, and its AP$_{75}$ value exceeds P2PFormer by 2.1, indicating that GCP generalizes better to unseen scenarios. This advantage may stem from GCP's mask-to-polygon pipeline, where mask predictions are usually less sensitive to domain shifts. Furthermore, the N-ratio values for GCP on both Test 1 and Test 2 are below 1.0, indicating that the generated polygons are more compact than the ground truth polygons.

\setlength\tabcolsep{0pt}
\begin{table*}
\footnotesize
\caption{
    Ablation study on WHU-Mix Vectors test1. ``Poly.'' denotes whether the binary masks predicted by Mask2Former are converted to polygons. $\mathcal{L}_{vert} + \mathcal{L}_{ang}$ denote whether the polyline regression head is used. $\mathcal{L}_{gc}$ denotes whether the global collinearity loss is employed. ``Simp.'' denotes which polygon simplification method is utlized. ``DP'' denotes the Douglas-Peucker algorithm. ``CLS'' denotes adding a classification head to filter out the redundant vertices. ``GCP'' denotes the proposed polyline simplification algorithm.
}
\renewcommand\arraystretch{1.2}
\newcolumntype{C}{>{\centering \arraybackslash}m{1.3cm}}
\newcolumntype{D}{>{\centering \arraybackslash}m{5.0cm}}
\newcolumntype{E}{>{\centering \arraybackslash}m{1.3cm}}
\begin{center}
    \begin{tabular}{E E E E | C C C C C C C C C C}
\toprule
        Poly. & $\mathcal{L}_{vert} + \mathcal{L}_{ang}$ & $\mathcal{L}_{gc}$ & Simp. & AP$\uparrow$ & AP$_{50}$$\uparrow$ & AP$_{75}$$\uparrow$ & IoU$\uparrow$ & MTA$\downarrow$ & C-IoU$\uparrow$ & N-ratio \\

\addlinespace[0.5ex]
\cline{1-11}
\addlinespace[0.5ex]
\ding{55} & - & - & - & \textbf{62.0} & \textbf{87.7} & \textbf{69.3} & \textbf{79.8} & - & - & - \\
\addlinespace[0.5ex]
\cline{1-11}
\addlinespace[0.5ex]
\checkmark & \ding{55} & \ding{55} & None & \textbf{62.0} & \textbf{87.7} & \textbf{69.3} & \textbf{79.8} & 37.2 & 12.5 & 13.7 \\
\checkmark & \ding{55} & \ding{55} & DP & 59.3 & 87.2 & 66.7 & 79.2 & 45.5 & 63.0 & 1.36 \\
\checkmark & \ding{55} & \ding{55} & GCP & 60.9 & 87.6 & 68.6 & 79.6 & 46.4 & 64.5 & 1.24 \\
\addlinespace[0.5ex]
\cline{1-11}
\addlinespace[0.5ex]
\checkmark & \checkmark & \ding{55} & CLS & 59.4 & 86.8 & 66.5 & 78.9 & 33.9 & 68.7 & 1.10 \\
\checkmark & \checkmark & \ding{55} & DP & 60.3 & 87.2 & 67.4 & 79.3 &  36.1 & 62.5 & 1.45\\
\checkmark & \checkmark & \ding{55} & GCP & 60.6 & 87.3 & 67.7 & 79.4 & 33.2 & 68.3 & 0.98   \\
\addlinespace[0.5ex]
\cline{1-11}
\addlinespace[0.5ex]
\checkmark & \checkmark & \checkmark & GCP & \textbf{62.0} & 87.6 & 69.1 & 79.6 & \textbf{32.4} & \textbf{68.8} & 0.96 \\

\addlinespace[0.5ex]
\bottomrule
\end{tabular}
\end{center}
\label{tab:ablation_study}
\end{table*}

\subsubsection{Ablation Study}
We conducted some ablation study experiments on the test set 1 of the WHU-Mix (vector) dataset.
As shown in Table \ref{tab:ablation_study}, we verify the influence of the polyline regression module, the proposed global collinearity loss $\mathcal{L}$, as well as the utilized polygon simplification method. 

The first row of Table \ref{tab:ablation_study} presents the performance of the predicted binary masks. The second row reflects the performance of a naive method that converts masks to polygons using the contour tracing algorithm \cite{suzuki1985topological}. While the accuracy of the predicted masks is largely preserved, this approach results in a significantly high N-ratio and very low C-IoU, indicating that the output polygons contain many redundant vertices.

The third and fourth rows denote the performance when simplifying the initial contour without using the polyline regression module. Here, we observe a drop in performance for the AP and IoU metrics, but a significant improvement in C-IoU values and a substantial reduction in N-ratio, suggesting that the polygons are significantly simplified.

In the last four rows, the polyline regression module is applied, utilizing the regression losses $\mathcal{L}_{vert}$ and $\mathcal{L}_{ang}$ to supervise the module. Comparing these results to those obtained with the polyline regression module disabled, we observe a significant reduction in MTA values. This indicates an increased geometric similarity between the predicted polygons and the ground truth polygons.

We also compare the effectiveness of different polygon simplification methods. In addition to the commonly used Douglas-Peucker algorithm (denoted as "DP") and the proposed GCP, we implement a simplification method referred to as "CLS," which classifies each point on the predicted contour as either a vertex or a non-vertex point. To achieve this, we add a vertex classification branch within the polyline regression head and filter out redundant points below a predefined threshold of 0.1 during the inference process.

Comparing these simplification methods, we find that "CLS" achieves lower MTA and N-ratio values than DP but results in a relatively significant performance decline in the AP, AP${50}$, and AP${75}$ metrics. Among the three methods, the proposed GCP consistently outperforms the others in terms of AP and IoU metrics, regardless of whether the polyline regression module is employed. Additionally, GCP achieves lower N-ratio and MTA values, indicating that it provides more compact and simplified results while effectively capturing the structural and geometric information of the building polygons.

The impact of the proposed $\mathcal{L}_{gc}$ loss can be observed by comparing the results of the last two rows. The inclusion of $\mathcal{L}_{gc}$ significantly enhances the overall performance of the polygonization pipeline. It not only reduces the performance gap between the polygonal and mask prediction results but also yields a more compact and simplified polygon representation.

In Fig. \ref{fig:ablation_study}, we present visualization results of several ablated models using different simplification methods on Test 1 and Test 2 of the WHU-Mix (vector) dataset. It can be observed that when the vertex classification-based method, ``CLS'', is used, some true building vertices may be missing, leading to the omission of significant portions of the building (as shown in the first row). When the Douglas-Peucker (``DP'') method is applied, certain redundant points on the building contour are not properly removed, resulting in an overly complex polygon representation. In contrast, the proposed GCP method is optimized to identify collinear relationships within the sampled building polylines, allowing it to generate a more compact and smooth polygon representation.

\setlength\tabcolsep{0pt}
\begin{table}
\footnotesize
\caption{
    Sensitivity of the proposed polyline simplification method to hyperparamter $\lambda$.
}
\renewcommand\arraystretch{1.2}
\newcolumntype{C}{>{\centering \arraybackslash}m{1.0cm}}
\newcolumntype{D}{>{\centering \arraybackslash}m{5.0cm}}
\newcolumntype{E}{>{\centering \arraybackslash}m{1.0cm}}
\begin{center}
    \begin{tabular}{E | C C C C C C C C C C}
\toprule
        $\lambda$ & AP & AP$_{50}$ & AP$_{75}$ & IoU & MTA & C-IoU & N-ratio \\

\addlinespace[0.5ex]
\cline{1-8}
\addlinespace[0.5ex]
 0 & 62.0 & 87.7 & 69.3 & 79.8 & 37.2 & 12.5 & 13.7  \\
 1 & 61.7 & 87.7 & 69.2 & 79.7 & 47.5 & 49.0 & 2.06 \\
 2 & 61.4 & 87.7 & 69.0 & 79.7 & 47.4 & 58.0 & 1.55 \\
 4 & 60.9 & 87.6 & 68.6 & 79.6 & 46.4 & 64.5 & 1.24 \\
 8 & 59.4 & 87.3 & 66.9 & 79.3 & 44.0 & 67.5 & 1.00 \\
\bottomrule
\end{tabular}
\end{center}
\label{tab:sensitivity}
\end{table}

\begin{figure*}[htp]
    \centering
    \includegraphics[width=0.99\linewidth]{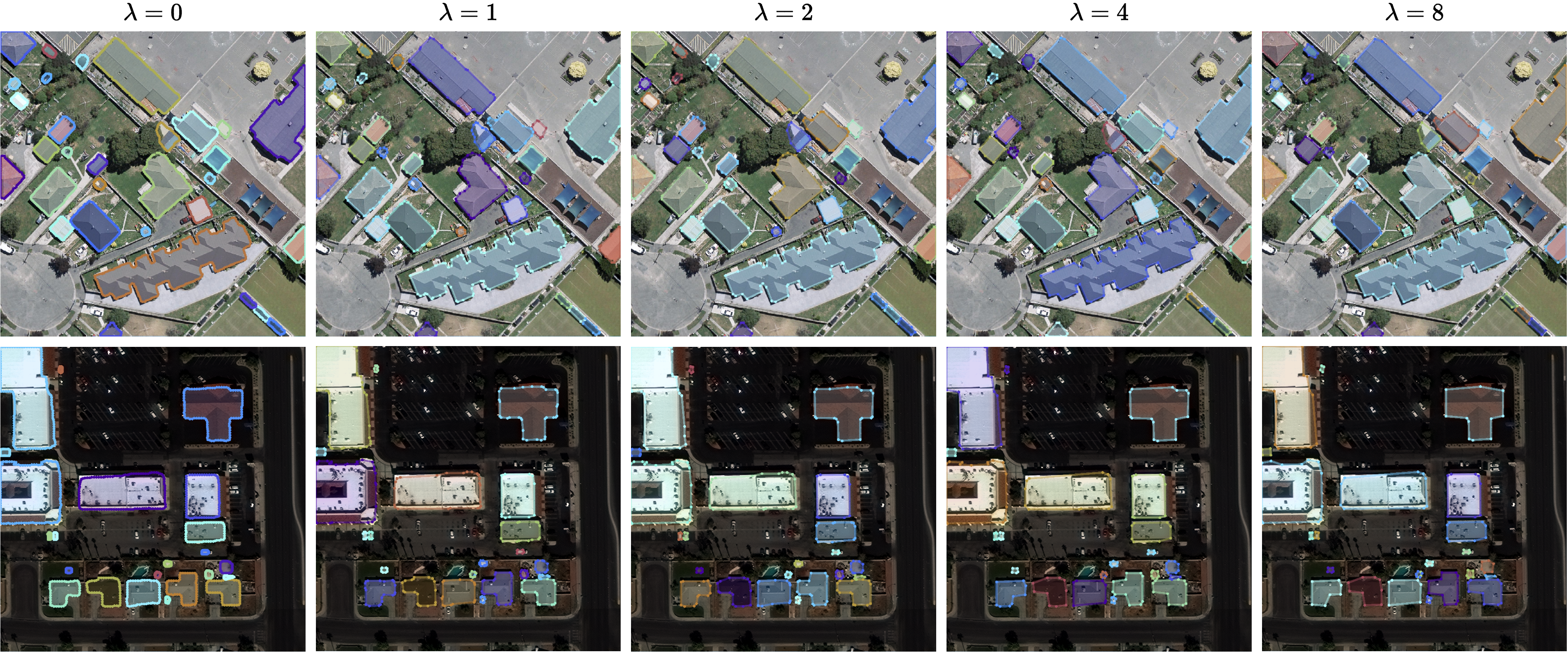}
    \caption{
Polygon simplification results of the proposed global collinearity-aware polyline simplification module under different $\lambda$ values.
The simplification is applied directly to initial contours sampled from building masks, without refinement from the polyline refinement module.
        }
    \label{fig:sensitivity}
\end{figure*}

\subsubsection{Sensitivity Analysis of $\lambda$}
\label{sec:sensitivity}
The only key hyperparameter in the proposed GCP approach is the $\lambda$ in Eq. \ref{eq:objective} that is used to balance the fidelity and complexity of the simplified polyline. 
Here we test the performance of the proposed polygon simplification method by using various $\lambda$ value to simplify the initial polygons sampled from the mask contours without being processed by the regression module.
According to Table \ref{tab:sensitivity}, we can observe that when $\lambda=0$, the simplification method deteriorate to naively keeping all the points in the initial polygon, resulting in a over complicated polygon representation.
With the increase of $\lambda$, the accuracy of the polygon gradually drops, resulting in lower AP and IoU values.
In the meanwhile, the complexity of the polygon results is gradually reduced, resulting in lower N-ratio and higher C-IoU values.

We visualize the simplification results for different $\lambda$ values in Fig. \ref{fig:sensitivity}. As the $\lambda$ value increases, the resulting polygons become progressively smoother.

\setlength\tabcolsep{0pt}
\begin{table}
\footnotesize
\caption{
    Computational Efficiency of the proposed GCP tested on Crowd AI dataset. The time consumption of each module to process a single $320 \times 320$ image are reported.
}
\renewcommand\arraystretch{1.2}
\newcolumntype{C}{>{\centering \arraybackslash}m{1.6cm}}
\newcolumntype{D}{>{\centering \arraybackslash}m{4.0cm}}
\begin{center}
    \begin{tabular}{ C C C C C }
\toprule
        Mask2Former & Poly OP & Poly Refine & Poly Simp. & Total  \\
\addlinespace[0.5ex]
\cline{1-5}
\addlinespace[0.5ex]
 4.9 ms & 17.4 ms & 4.6 ms & 5.3 ms & 32.3 ms \\
        
\bottomrule
\end{tabular}
\end{center}
\label{tab:time_consumption}
\end{table}

\subsubsection{Computational Efficiency}
We report the time consumption of different modules of the proposed GCP on the CrowdAI dataset. In Table \ref{tab:time_consumption}, ``Mask2Former'' refers to the time required for Mask2Former to generate building masks. "Poly OP" represents polygon-related operations such as converting binary masks to initial polygons, sampling points along the polygon contour, assembling the refined polylines into closed polygons, and other similar processes. All these operations are executed on a single CPU core. ``Poly Refine'' indicates the time consumption of the polyline regression module, while ``Poly Simp.'' refers to the proposed polygon simplification method.

Overall, it can be observed that mask generation, polyline refinement, and polygon simplification each consume approximately the same amount of time. Other auxiliary polygon operations take more time but can be easily accelerated using multi-processing techniques for time-sensitive applications.

\subsection{Failure Cases}
\label{sec:failure}
\begin{figure}[htp]
    \centering
    \includegraphics[width=1.0\linewidth]{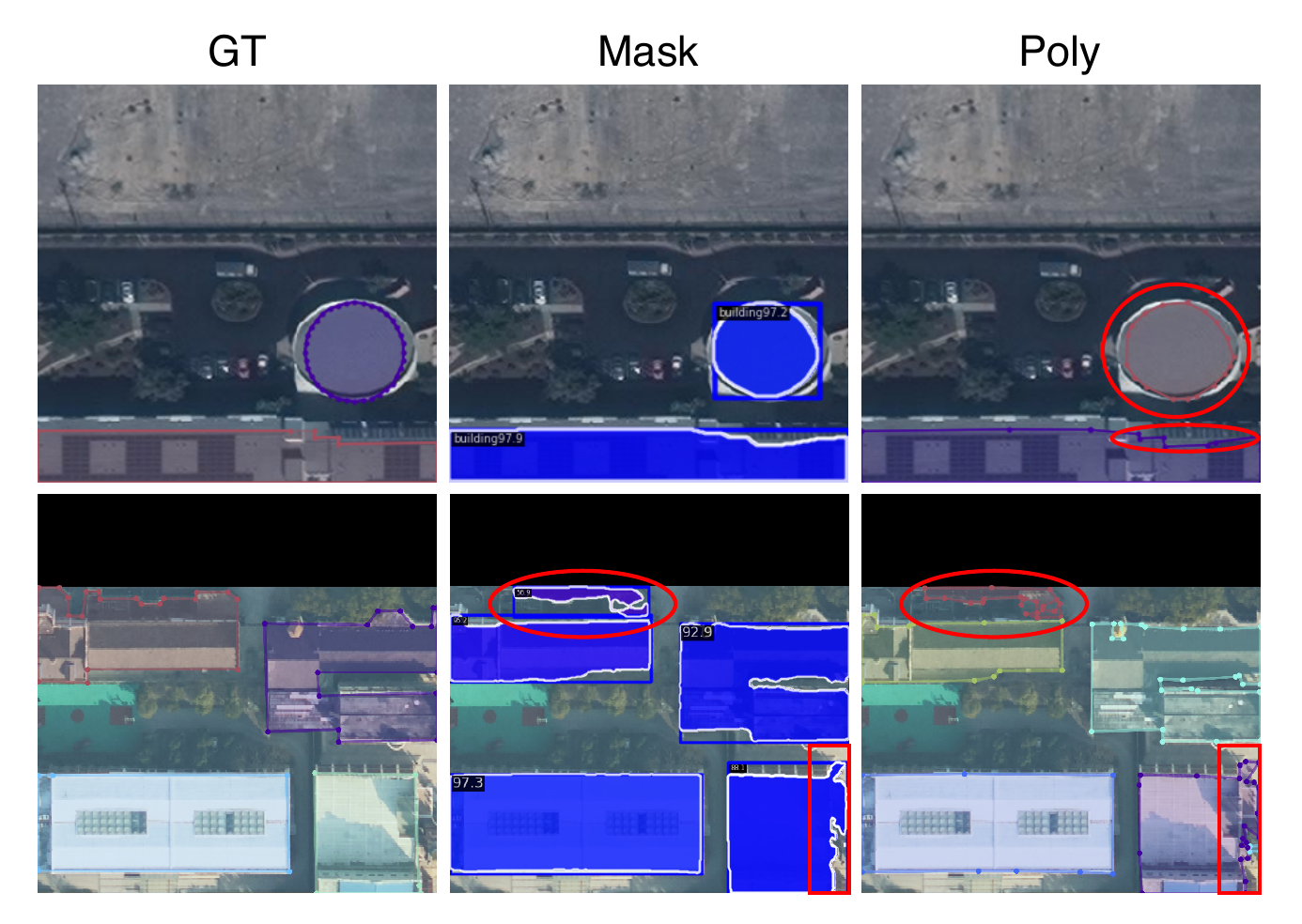}
    \caption{
Failure cases of the proposed GCP method. The first row is from the CrowdAI dataset, while the second row is from the WHU-Mix (vector) dataset.
    }
    \label{fig:failure_cases}
\end{figure}

We present two typical failure cases of the proposed GCP method on the CrowdAI and WHU-Mix (vector) datasets in Fig. \ref{fig:failure_cases}.
In the first row, GCP struggles with rounded building structures, as it tends to approximate building geometry using a piecewise linear representation, resulting in fewer vertices and less accurate outlines.
The second row illustrates that the predicted building polygons are dependent on the predicted binary building masks. If the masks fail to capture the rough building geometries accurately, the polygon extraction process is also likely to fail.

A potential solution for better polygonization of rounded buildings is to adaptively adjust the regularization coefficient $\lambda$ based on building type, allowing more vertices for curved structures. For the second issue, GCP’s performance could be improved by integrating a more powerful instance segmentation model.

%% file: content/conclusion.tex
In this paper, we propose a novel polygonal building mapping method called GCP, aimed at addressing the challenges of converting building mask representations into precise polygon representations. We introduce a polyline refinement module to enhance initial polygons sampled along the boundaries of building masks. This is followed by a global collinearity-based polyline simplification module, which processes the refined polylines to generate the final predicted polygons. Additionally, we incorporate a global collinearity loss that is differentiable with respect to the refined polylines, enabling the entire polygonization pipeline to be trained end-to-end.

Extensive experiments demonstrate that the proposed method delivers superior performance. Furthermore, the polyline simplification module, when applied independently, outperforms traditional methods such as the Douglas-Peucker algorithm, extending its potential for broader applications.

%% file: bare_jrnl.bbl
\begin{thebibliography}{10}
\providecommand{\url}[1]{#1}
\csname url@samestyle\endcsname
\providecommand{\newblock}{\relax}
\providecommand{\bibinfo}[2]{#2}
\providecommand{\BIBentrySTDinterwordspacing}{\spaceskip=0pt\relax}
\providecommand{\BIBentryALTinterwordstretchfactor}{4}
\providecommand{\BIBentryALTinterwordspacing}{\spaceskip=\fontdimen2\font plus
\BIBentryALTinterwordstretchfactor\fontdimen3\font minus \fontdimen4\font\relax}
\providecommand{\BIBforeignlanguage}[2]{{%
\expandafter\ifx\csname l@#1\endcsname\relax
\typeout{** WARNING: IEEEtran.bst: No hyphenation pattern has been}%
\typeout{** loaded for the language `#1'. Using the pattern for}%
\typeout{** the default language instead.}%
\else
\language=\csname l@#1\endcsname
\fi
#2}}
\providecommand{\BIBdecl}{\relax}
\BIBdecl

\bibitem{zhu2024global}
X.~X. Zhu, Q.~Li, Y.~Shi, Y.~Wang, A.~Stewart, and J.~Prexl, ``Global openbuildingmap--unveiling the mystery of global buildings,'' \emph{arXiv preprint arXiv:2404.13911}, 2024.

\bibitem{robinson2017deep}
C.~Robinson, F.~Hohman, and B.~Dilkina, ``A deep learning approach for population estimation from satellite imagery,'' in \emph{Proceedings of the 1st ACM SIGSPATIAL Workshop on Geospatial Humanities}, 2017, pp. 47--54.

\bibitem{rentschler2022flood}
J.~Rentschler, M.~Salhab, and B.~A. Jafino, ``Flood exposure and poverty in 188 countries,'' \emph{Nature communications}, vol.~13, no.~1, p. 3527, 2022.

\bibitem{guo2024high}
J.~Guo, D.~Hong, and X.~X. Zhu, ``High-resolution satellite images reveal the prevalent positive indirect impact of urbanization on urban tree canopy coverage in south america,'' \emph{Landscape and Urban Planning}, vol. 247, p. 105076, 2024.

\bibitem{jiang2021industrial}
Y.~Jiang, S.~Yin, K.~Li, H.~Luo, and O.~Kaynak, ``Industrial applications of digital twins,'' \emph{Philosophical Transactions of the Royal Society A}, vol. 379, no. 2207, p. 20200360, 2021.

\bibitem{chen2021mask}
S.~Chen, L.~Mou, Q.~Li, Y.~Sun, and X.~X. Zhu, ``Mask-height r-cnn: An end-to-end network for 3d building reconstruction from monocular remote sensing imagery,'' in \emph{2021 IEEE International Geoscience and Remote Sensing Symposium IGARSS}.\hskip 1em plus 0.5em minus 0.4em\relax IEEE, 2021, pp. 1202--1205.

\bibitem{xu2023hisup}
B.~Xu, J.~Xu, N.~Xue, and G.-S. Xia, ``Hisup: Accurate polygonal mapping of buildings in satellite imagery with hierarchical supervision,'' \emph{ISPRS Journal of Photogrammetry and Remote Sensing}, vol. 198, pp. 284--296, 2023.

\bibitem{girard2021polygonal}
N.~Girard, D.~Smirnov, J.~Solomon, and Y.~Tarabalka, ``Polygonal building extraction by frame field learning,'' in \emph{Proceedings of the IEEE/CVF Conference on Computer Vision and Pattern Recognition}, 2021, pp. 5891--5900.

\bibitem{lorensen1998marching}
W.~E. Lorensen and H.~E. Cline, ``Marching cubes: A high resolution 3d surface construction algorithm,'' in \emph{Seminal graphics: pioneering efforts that shaped the field}, 1998, pp. 347--353.

\bibitem{suzuki1985topological}
S.~Suzuki \emph{et~al.}, ``Topological structural analysis of digitized binary images by border following,'' \emph{Computer vision, graphics, and image processing}, vol.~30, no.~1, pp. 32--46, 1985.

\bibitem{douglas1973algorithms}
D.~H. Douglas and T.~K. Peucker, ``Algorithms for the reduction of the number of points required to represent a digitized line or its caricature,'' \emph{Cartographica: the international journal for geographic information and geovisualization}, vol.~10, no.~2, pp. 112--122, 1973.

\bibitem{krizhevsky2012imagenet}
A.~Krizhevsky, I.~Sutskever, and G.~E. Hinton, ``Imagenet classification with deep convolutional neural networks,'' \emph{Advances in neural information processing systems}, vol.~25, 2012.

\bibitem{hu2023polybuilding}
Y.~Hu, Z.~Wang, Z.~Huang, and Y.~Liu, ``Polybuilding: Polygon transformer for building extraction,'' \emph{ISPRS Journal of Photogrammetry and Remote Sensing}, vol. 199, pp. 15--27, 2023.

\bibitem{zhang2024hit}
M.~Zhang, Q.~Liu, and Y.~Wang, ``Hit: Building mapping with hierarchical transformers,'' \emph{IEEE Transactions on Geoscience and Remote Sensing}, 2024.

\bibitem{carion2020end}
N.~Carion, F.~Massa, G.~Synnaeve, N.~Usunier, A.~Kirillov, and S.~Zagoruyko, ``End-to-end object detection with transformers,'' in \emph{European conference on computer vision}.\hskip 1em plus 0.5em minus 0.4em\relax Springer, 2020, pp. 213--229.

\bibitem{cuturi2013sinkhorn}
M.~Cuturi, ``Sinkhorn distances: Lightspeed computation of optimal transport,'' \emph{Advances in neural information processing systems}, vol.~26, 2013.

\bibitem{zorzi2022polyworld}
S.~Zorzi, S.~Bazrafkan, S.~Habenschuss, and F.~Fraundorfer, ``Polyworld: Polygonal building extraction with graph neural networks in satellite images,'' in \emph{Proceedings of the IEEE/CVF Conference on Computer Vision and Pattern Recognition}, 2022, pp. 1848--1857.

\bibitem{bellman1966dynamic}
R.~Bellman, ``Dynamic programming,'' \emph{science}, vol. 153, no. 3731, pp. 34--37, 1966.

\bibitem{zhao2018building}
K.~Zhao, J.~Kang, J.~Jung, and G.~Sohn, ``Building extraction from satellite images using mask r-cnn with building boundary regularization,'' in \emph{Proceedings of the IEEE conference on computer vision and pattern recognition workshops}, 2018, pp. 247--251.

\bibitem{he2017mask}
K.~He, G.~Gkioxari, P.~Doll{\'a}r, and R.~Girshick, ``Mask r-cnn,'' in \emph{Proceedings of the IEEE international conference on computer vision}, 2017, pp. 2961--2969.

\bibitem{wei2019toward}
S.~Wei, S.~Ji, and M.~Lu, ``Toward automatic building footprint delineation from aerial images using cnn and regularization,'' \emph{IEEE Transactions on Geoscience and Remote Sensing}, vol.~58, no.~3, pp. 2178--2189, 2019.

\bibitem{li2020approximating}
M.~Li, F.~Lafarge, and R.~Marlet, ``Approximating shapes in images with low-complexity polygons,'' in \emph{Proceedings of the IEEE/CVF Conference on Computer Vision and Pattern Recognition}, 2020, pp. 8633--8641.

\bibitem{bauchet2018kippi}
J.-P. Bauchet and F.~Lafarge, ``Kippi: Kinetic polygonal partitioning of images,'' in \emph{Proceedings of the IEEE conference on computer vision and pattern recognition}, 2018, pp. 3146--3154.

\bibitem{kass1988snakes}
M.~Kass, A.~Witkin, and D.~Terzopoulos, ``Snakes: Active contour models,'' \emph{International journal of computer vision}, vol.~1, no.~4, pp. 321--331, 1988.

\bibitem{wei2023buildmapper}
S.~Wei, T.~Zhang, S.~Ji, M.~Luo, and J.~Gong, ``Buildmapper: A fully learnable framework for vectorized building contour extraction,'' \emph{ISPRS Journal of Photogrammetry and Remote Sensing}, vol. 197, pp. 87--104, 2023.

\bibitem{zorzi2023re}
S.~Zorzi and F.~Fraundorfer, ``Re: Polyworld-a graph neural network for polygonal scene parsing,'' in \emph{Proceedings of the IEEE/CVF International Conference on Computer Vision}, 2023, pp. 16\,762--16\,771.

\bibitem{wei2024lines}
S.~Wei, T.~Zhang, D.~Yu, S.~Ji, Y.~Zhang, and J.~Gong, ``From lines to polygons: Polygonal building contour extraction from high-resolution remote sensing imagery,'' \emph{ISPRS Journal of Photogrammetry and Remote Sensing}, vol. 209, pp. 213--232, 2024.

\bibitem{hosang2017learning}
J.~Hosang, R.~Benenson, and B.~Schiele, ``Learning non-maximum suppression,'' in \emph{Proceedings of the IEEE conference on computer vision and pattern recognition}, 2017, pp. 4507--4515.

\bibitem{medsker2001recurrent}
L.~R. Medsker, L.~Jain \emph{et~al.}, ``Recurrent neural networks,'' \emph{Design and Applications}, vol.~5, no. 64-67, p.~2, 2001.

\bibitem{vaswani2017attention}
A.~Vaswani, ``Attention is all you need,'' \emph{Advances in Neural Information Processing Systems}, 2017.

\bibitem{li2019topological}
Z.~Li, J.~D. Wegner, and A.~Lucchi, ``Topological map extraction from overhead images,'' in \emph{Proceedings of the IEEE/CVF International Conference on Computer Vision}, 2019, pp. 1715--1724.

\bibitem{huang2021oec}
W.~Huang, H.~Tang, and P.~Xu, ``Oec-rnn: Object-oriented delineation of rooftops with edges and corners using the recurrent neural network from the aerial images,'' \emph{IEEE Transactions on Geoscience and Remote Sensing}, vol.~60, pp. 1--12, 2021.

\bibitem{shi2015convolutional}
X.~Shi, Z.~Chen, H.~Wang, D.-Y. Yeung, W.-K. Wong, and W.-c. Woo, ``Convolutional lstm network: A machine learning approach for precipitation nowcasting,'' \emph{Advances in neural information processing systems}, vol.~28, 2015.

\bibitem{zhang2024p2pformer}
T.~Zhang, S.~Wei, Y.~Zhou, M.~Luo, W.~Yu, and S.~Ji, ``P2pformer: A primitive-to-polygon method for regular building contour extraction from remote sensing images,'' \emph{IEEE Transactions on Geoscience and Remote Sensing}, 2024.

\bibitem{ren2016faster}
S.~Ren, K.~He, R.~Girshick, and J.~Sun, ``Faster r-cnn: Towards real-time object detection with region proposal networks,'' \emph{IEEE transactions on pattern analysis and machine intelligence}, vol.~39, no.~6, pp. 1137--1149, 2016.

\bibitem{cheng2021per}
B.~Cheng, A.~Schwing, and A.~Kirillov, ``Per-pixel classification is not all you need for semantic segmentation,'' \emph{Advances in neural information processing systems}, vol.~34, pp. 17\,864--17\,875, 2021.

\bibitem{cheng2022masked}
B.~Cheng, I.~Misra, A.~G. Schwing, A.~Kirillov, and R.~Girdhar, ``Masked-attention mask transformer for universal image segmentation,'' in \emph{Proceedings of the IEEE/CVF conference on computer vision and pattern recognition}, 2022, pp. 1290--1299.

\bibitem{li2022mask}
F.~Li, H.~Zhang, H.~xu, S.~Liu, L.~Zhang, L.~M. Ni, and H.-Y. Shum, ``Mask dino: Towards a unified transformer-based framework for object detection and segmentation,'' 2022.

\bibitem{kuhn1955hungarian}
H.~W. Kuhn, ``The hungarian method for the assignment problem,'' \emph{Naval research logistics quarterly}, vol.~2, no. 1-2, pp. 83--97, 1955.

\bibitem{crowdai2023mapping}
crowdAI, ``Mapping challenge,'' \url{https://www.crowdai.org/challenges/mapping-challenge}, 2023, accessed: 2023-09-28.

\bibitem{adimoolam2023efficient}
Y.~K. Adimoolam, B.~Chatterjee, C.~Poullis, and M.~Averkiou, ``Efficient deduplication and leakage detection in large scale image datasets with a focus on the crowdai mapping challenge dataset,'' \emph{arXiv preprint arXiv:2304.02296}, 2023.

\bibitem{ji2018fully}
S.~Ji, S.~Wei, and M.~Lu, ``Fully convolutional networks for multisource building extraction from an open aerial and satellite imagery data set,'' \emph{IEEE Transactions on geoscience and remote sensing}, vol.~57, no.~1, pp. 574--586, 2018.

\bibitem{codalab20100}
{OpenAI}, ``2018 open ai tanzania building footprint segmentation challenge,'' \url{https://competitions.codalab.org/competitions/20100}, 2023, accessed: 2023-09-28.

\bibitem{van2018spacenet}
A.~Van~Etten, D.~Lindenbaum, and T.~M. Bacastow, ``Spacenet: A remote sensing dataset and challenge series,'' \emph{arXiv preprint arXiv:1807.01232}, 2018.

\bibitem{maggiori2017can}
E.~Maggiori, Y.~Tarabalka, G.~Charpiat, and P.~Alliez, ``Can semantic labeling methods generalize to any city? the inria aerial image labeling benchmark,'' in \emph{2017 IEEE International geoscience and remote sensing symposium (IGARSS)}.\hskip 1em plus 0.5em minus 0.4em\relax IEEE, 2017, pp. 3226--3229.

\bibitem{he2016deep}
K.~He, X.~Zhang, S.~Ren, and J.~Sun, ``Deep residual learning for image recognition,'' in \emph{Proceedings of the IEEE conference on computer vision and pattern recognition}, 2016, pp. 770--778.

\bibitem{loshchilov2017decoupled}
I.~Loshchilov, ``Decoupled weight decay regularization,'' \emph{arXiv preprint arXiv:1711.05101}, 2017.

\end{thebibliography}
